\documentclass[letterpaper, 10 pt, journal, twoside]{IEEEtran}

\usepackage{amsmath,amsfonts}
\usepackage{algorithmic}
\usepackage{algorithm}
\usepackage{array}
\usepackage{textcomp}
\usepackage{stfloats}
\usepackage{url}
\usepackage{verbatim}
\usepackage{graphicx}
\usepackage{cite}
\usepackage{booktabs}
\usepackage{amssymb}
\usepackage{ftnxtra}
\usepackage{subcaption}
\usepackage{wrapfig}
\usepackage{multirow}

\usepackage[accsupp]{axessibility}  

\usepackage[pagebackref,breaklinks,colorlinks]{hyperref}

\usepackage[capitalize]{cleveref}
\crefname{section}{Sec.}{Secs.}
\Crefname{section}{Section}{Sections}
\Crefname{table}{Table}{Tables}
\crefname{table}{Tab.}{Tabs.}
\usepackage{multirow}

\usepackage{orcidlink}

\author{
Yue Hu, Juntong Peng, Yunqiao Yang, Siheng Chen~\IEEEmembership{Senior Member}
\thanks{All authors are with Shanghai Jiao Tong University, Shanghai, China (e-mail: 18671129361, juntong.peng, yangyunqiao, sihengc@sjtu.edu.cn). }
}

\begin{document}

\title{Communication-Efficient Multi-Agent 3D Detection via Hybrid Collaboration}

\maketitle



\begin{abstract}

Collaborative 3D detection can substantially boost detection performance by allowing agents to exchange complementary information. It inherently results in a fundamental trade-off between detection performance and communication bandwidth.
To tackle this bottleneck issue, we propose a novel hybrid collaboration that adaptively integrates two types of communication messages: perceptual outputs, which are compact, and raw observations, which offer richer information. 
This approach focuses on two key aspects: i) integrating complementary information from two message types and ii) prioritizing the most critical data within each type. By adaptively selecting the most critical set of messages, it ensures optimal perceptual information and adaptability, effectively meeting the demands of diverse communication scenarios.
Building on this hybrid collaboration, we present \texttt{HyComm}, a communication-efficient LiDAR-based collaborative 3D detection system. \texttt{HyComm} boasts two main benefits: i) it facilitates adaptable compression rates for messages, addressing various communication requirements, and ii) it uses standardized data formats for messages. This ensures they are independent of specific detection models, fostering adaptability across different agent configurations. 
To evaluate~\texttt{HyComm}, we conduct experiments on both real-world and simulation datasets: DAIR-V2X and OPV2V. \texttt{HyComm} consistently outperforms previous methods and achieves a superior performance-bandwidth trade-off regardless of whether agents use the same or varied detection models. It achieves a lower communication volume of more than 2,006$\times$ and still outperforms Where2comm on DAIR-V2X in terms of AP50. The related code will be released.

%

\end{abstract}
\vspace{-2mm}
\vspace{-3mm}

\section{Introduction}
\label{sec:intro}

\IEEEPARstart{C}{ollaborative} perception targets to improve perception performance by enabling agents to exchange complementary information.
It can fundamentally overcome the occlusion and long-range issues in single-agent perception due to the limited visibility from a single view. 
To achieve collaborative perception, previous works have contributed high-quality datasets~\cite{V2XSim,OPV2V,dair,hu2022where2comm} and effective collaboration methods~\cite{zhou2022multi,ArnoldDT:22,v2vnet,Li2022mrsc,SuUncertainty:ICRA23,CoBEVT} to address the numerous issues in this emerging field, such as communication bandwidth constraints~\cite{when2com,who2com,disconet,hu2022where2comm,yuan2022keypoints,HuCollaboration:CVPR23}, latency~\cite{SyncNet,xu2022v2x}, and pose error~\cite{PoseError,lu2022robust}. This work addresses the bottleneck in scaling up collaborative perception: the communication bandwidth constraint challenge. As in real-world scenarios, communication resources are always limited, and inefficient use of bandwidth restricts collaborative perception to a small number of agents and a limited timeframe. This limitation results in only marginal performance improvements over single-agent perception.

\begin{figure}[!t]
\centering
    \includegraphics[width=1.0\linewidth]{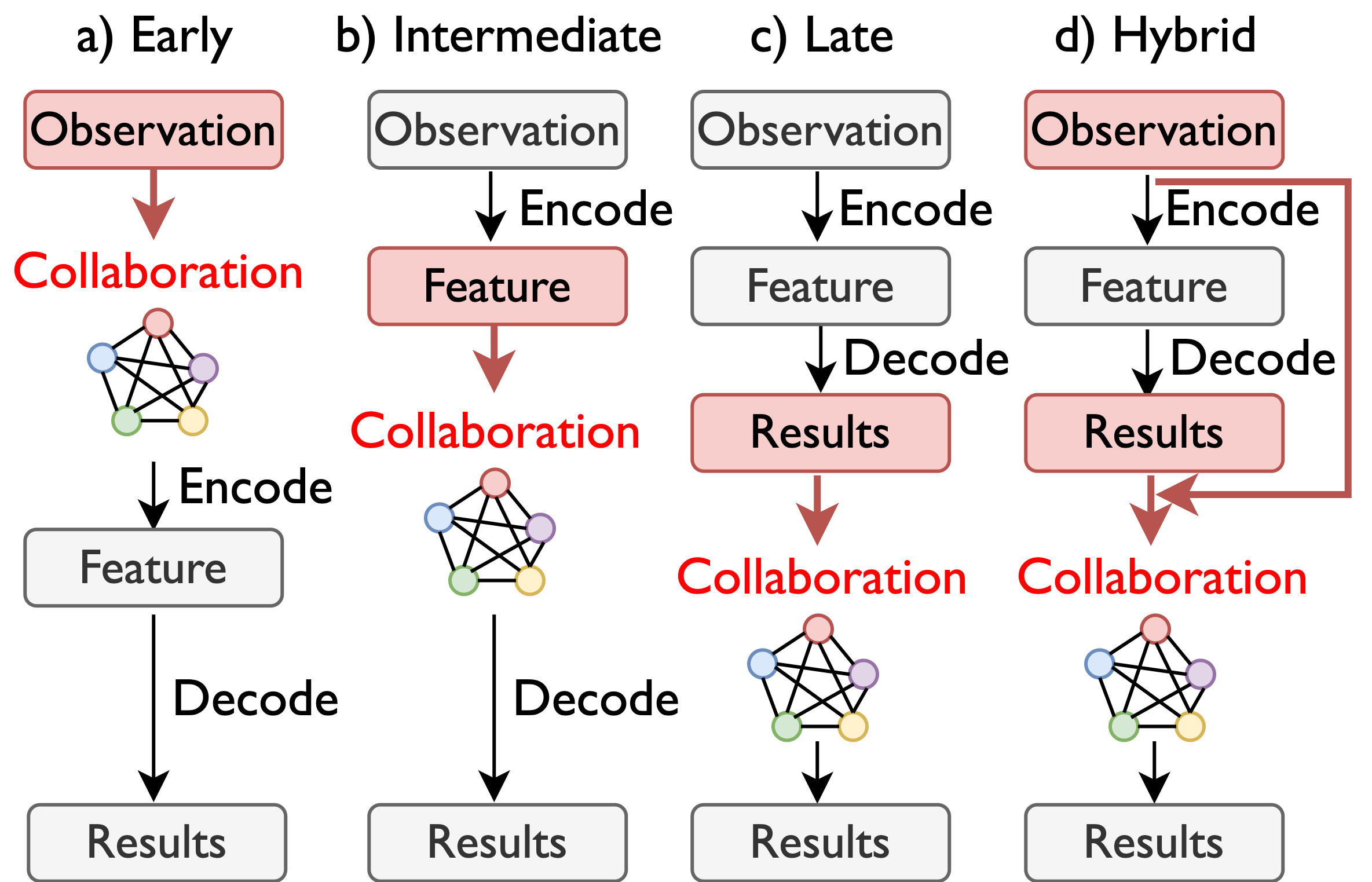}
    \vspace{-3mm}
    \caption{Our novel hybrid collaboration exchanges both point and box among agents. It allows for flexibly adjusting compression rates to adapt to the entire communication spectrum.}
    \label{Fig:intro}
    \vspace{-6mm}
\end{figure}

To optimize communication-constrained collaborative perception, previous works consider three types of collaboration strategies: i) early collaboration~\cite{Chen2019CooperCP}, whose transmitted message is raw observations, fitting for situations with ample communication budget; ii) intermediate collaboration~\cite{when2com,who2com,hu2022where2comm,disconet,V2XSim,xu2022v2x,dair,CoBEVT,HuCollaboration:CVPR23,YangWhat2comm:ICCV2023,How2comm:NeurIPS2023}, whose transmitted message is hidden features, fitting for situations with moderate communication budget; and iii) late collaboration, which transmits perceptual output, fitting for situations with highly restricted communication budget. However, each communication message type excels only within a narrow range of communication budget and lacks the adaptability to handle varying communication budgets.

To overcome this limitation, a hybrid collaboration strategy that integrates multiple collaboration types presents a highly adaptable and promising solution. It can take advantage of the strengths of diverse collaboration messages. For instance, combining the compact output-based messages with the comprehensive raw-observation-based messages, the hybrid message can potentially enable adaptive compression rates to accommodate varying communication budgets. However, this approach remains largely underexplored. The existing hybrid collaboration method~\cite{ArnoldDT:22} relies on a heuristic switch between late and early collaboration, lacking adaptability and failing to achieve optimal performance.

To address this issue, we propose a novel hybrid collaboration strategy that adaptively integrates late and early collaboration, ensuring optimal perceptual information. This strategy focuses on two key aspects. First, it leverages uncertainty as guidance to adaptively blend two message types: raw observation and perceptual output. Uncertainty reflects the inaccuracy level of the perceptual output. Guided by this, accurately perceived instances are conveyed in the most compact form, the output, while less reliable ones are selectively supplemented with additional raw observations. This adaptive integration ensures that perceptual information is initially conveyed in the most compact form, with the most beneficial details incrementally supplemented, ensuring an optimal composition of perceptual information from two message types. 
Second, it prioritizes the most important messages within each message type. Each perceptual output or raw observation is assigned an importance score indicating its perceptual significance. Such prioritization guarantees optimal perceptual information relay for each message type.

Following this novel hybrid collaboration strategy, we propose \texttt{HyComm}, a novel communication-efficient multi-agent collaborative LiDAR-based 3D object detection system. 
It includes four key modules: i) single-agent detector with confidence and uncertainty estimation, reflecting the detection confidence and uncertainty. These metrics decide the perceptual significance of each message type and guide the selection of the hybrid message; ii) a novel confidence-based box message packer, which prioritizes the boxes with higher confidence, considering them to offer optimal perceptual information in the most compact form; iii) a novel uncertainty-based point message packer, which emphasizes points from instances with larger uncertainty, considering them to offer optimal complementary perceptual information to address uncertainty; and iv) fusion module, which integrates the hybrid messages to achieve enhanced collaborative detections.



\texttt{HyComm} has three distinct benefits: i) compared with early collaboration and late collaboration, it offers flexible compression rates to cater to specific bandwidth requirements and substantially boost performance-bandwidth trade-offs; ii) compared with intermediate collaboration, the hybrid message adopts standardized box and point data formats that are independent of specific detection models, allowing collaboration among agents with diverse setups. This promotes a compatible collaboration system; and iii) compare with previous hybrid collaboration method~\cite{ArnoldDT:22}, which heuristically switches between early and late collaboration and transmit all the data in each collaboration mode, it can adaptively supplement inaccurate boxes with more precise raw point data, effectively mitigating the potential errors in late collaboration with early collaboration, ensuring robust performance. 

We conduct extensive experiments on both simulation and real-world datasets, OPV2V and DAIR-V2X. The results show that i) \texttt{HyComm} achieves superior performance-communication trade-off, outperforming intermediate collaboration with 2,006/1,317 $\times$ less communication cost on DAIR-V2X/OPV2V; ii) \texttt{HyComm} maintains superior performance-communication trade-off under the heterogeneous situation that agents are equipped with different models, while intermediate collaboration methods get a lower performance than single agent perception due to the misaligned feature maps generated by different models; and iii) both the confidence and uncertainty-guided hybrid message packers are beneficial.

To sum up, our main contributions are three-fold:

$\bullet$ We propose a novel hybrid collaboration strategy with improved adaptability, which substantially enhances performance-bandwidth trade-offs across the entire range of communication constraints;

$\bullet$ We propose \texttt{HyComm}, a novel communication-efficient collaboration system, which achieves efficient hybrid collaboration with the proposed confidence/uncertainty-based box/point message packer;

$\bullet$ We conduct extensive experiments to validate that \texttt{HyComm} achieves superior performance-communication trade-off across various agent setups, whether using identical or different detection models.

\vspace{-5mm}
\section{Related works}
\label{sec:related_work}
\vspace{-1mm}
\subsection{Communication efficiency in collaborative perception}
\vspace{-1mm}
Communication efficiency is the bottleneck for the scale-up of collaborative perception. The communication cost increases linearly with the size of perceptual regions and quadratically with the number of collaborating agents. Real-world communication resource is always constrained and can hardly support huge communication consumption in real-time.
There are three main collaboration messages types: i) early collaboration~\cite{Chen2019CooperCP}, which transmits raw observation data, suitable for contexts with a generous communication budget; ii) intermediate collaboration~\cite{hu2022where2comm,when2com,v2vnet,disconet,xu2022v2x,dair,CoBEVT,HuCollaboration:CVPR23,YangWhat2comm:ICCV2023,WangUMC:ICCV23,How2comm:NeurIPS2023,YangSpatioTemporalDA:ICCV2023,XuBridging:ICRA23}, which transmits representative information with compact features, ideal for scenarios with a medium-scale communication budget; and iii) late collaboration, which transmits detected boxes, tailored for situations where the communication budget is highly constrained. 
Aiming to harness the strengths of these collaboration message types, we introduce a novel uncertainty-guided hybrid collaboration. It uses uncertainty as a guidance to adaptively integrate efficient late collaboration and informative early collaboration, offering an effective communication solution for the varying communication budgets. 
 \vspace{-5mm}
\subsection{Uncertainty estimation}
Uncertainty estimation for the neural network assigns a distribution (by adding a variance parameter) over the weights of the neural network to estimate epistemic uncertainty~\cite{Neal1995BayesianLF}. Besides adding distribution directly to weights, the dropout layer can be leveraged~\cite{Kendall2017WhatUD} as an alternative way to estimate the final posterior and reflect the uncertainty.
In object detection, uncertainty estimation focuses on capturing the regression variance of detection results. The Von-Mises NLL loss function is proposed in ~\cite{UncertaintyLoss} to promote the uncertainty estimation accuracy, and demonstrated to assign large uncertainty values to inaccurate detections, thereby improving detection performance in the 3D object detector SECOND~\cite{SECOND}.
Here, we leverage object detection with uncertainty estimation to identify unreliable box messages, those detections with higher uncertainty scores. This identification allows our hybrid collaboration to select additional point messages for these inaccurate boxes, promising more precise detection results with the complementary point clouds.


\begin{figure*}[!t]
    \includegraphics[width=1.0\linewidth]{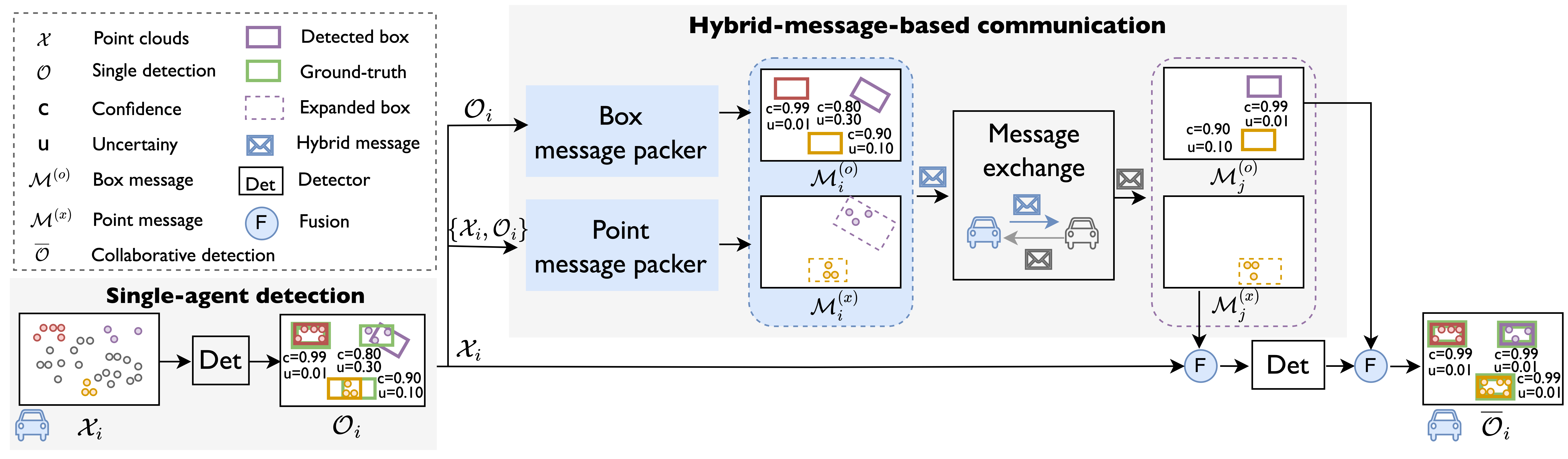}
    \vspace{-6mm}
    \caption{System overview. In \texttt{HyComm}, single-agent detection generates boxes with confidence and uncertainty. These detections, in conjunction with raw point clouds, are processed by the box message packer and point message packer, resulting in box messages and point messages. The hybrid messages are then exchanged among agents. Once received, these messages are fused to produce improved collaborative detections.}
    \label{Fig:framework}
    \vspace{-6mm}
\end{figure*}



\vspace{-4mm}
\section{Hybrid Collaborative 3D Detection}
\label{sec:method}
\vspace{-1mm}

This section presents \texttt{HyComm}, a communication-efficient multi-agent perception framework based on a novel hybrid collaboration strategy; see Fig.~\ref{Fig:framework}. Here, agents indicate intelligent vehicles and infrastructures that can communicate and share information in transportation scenarios. The overall architecture is defined in Sec.~\ref{subsec:4} and implementations are introduced in the following Sec.~\ref{subsec:4a},~\ref{subsec:4b},~\ref{subsec:4c},~\ref{subsec:4d}. 

\vspace{-4mm}
\subsection{Overall architecture}\label{subsec:4}
\vspace{-1mm}

Consider $N$ agents in the scene, each agent has the ability to perceive, communicate, and detect. Collaborative perception allows these agents to exchange complementary perceptual information, thereby improving overall perception performance.  To adjust to the dynamic and limited communication resources in real-world scenarios, the objective of collaborative perception is to consistently maximize the perception performance while adapting to varying communication bandwidth budgets. 
Different from previous collaboration strategies that rely on a single message type to manage the varying communication conditions, we propose a novel hybrid collaboration, which combines compact box messages with informative point messages. In doing so, it amalgamates the efficiency of late collaboration under limited budget constraints with the effectiveness of early collaboration when budget is abundant, offering a solution that adapts to the entire communication budget range.
To elaborate further, it is accomplished by giving precedence to efficient box messages and augmenting them with supplementary information from point messages as the communication budget expands.

Let $\mathcal{X}_i$, $\mathcal{O}_i$, $B$ be the raw point clouds, individual detection results, and communication budget respectively. The collaborative 3D detection based on hybrid collaboration is 
\begin{subequations}
\vspace{-2mm}
\label{eq:formulation}
\begin{align}
    \mathcal{O}_i &= \Phi_{\rm detect}(\mathcal{X}_i), \label{subeq:detection}\\ 
    \mathcal{M}_i^{(o)} &=\Phi_{\rm box}(\mathcal{O}_i, B),\label{subeq:late}\\
    \mathcal{M}_i^{(x)} &=\Phi_{\rm point}(\mathcal{X}_i, \mathcal{O}_i, B),\label{subeq:early}\\
    \mathcal{M}_i &= \{\mathcal{M}_i^{(x)}, \mathcal{M}_i^{(o)}\}, \label{subeq:message}\\ 
    \overline{\mathcal{O}}_i &= \Phi_{\rm fusion}(\mathcal{X}_i,\{\mathcal{M}_j\}_{j\in\mathcal{N}_i}),
    \label{subeq:fusion}
\end{align}
\end{subequations}
where $\Phi_{\rm detect}(\cdot)$ is the single-agent detector,
$\mathcal{M}_i$ represents the hybrid collaboration message from the $i$th agent, comprising two elements: the box message, $\mathcal{M}_i^{(o)}$, and the point message, $\mathcal{M}_i^{(x)}$. The neighbor agent set of the $i$th agent is denoted as $\mathcal{N}_i$,
$\Phi_{\rm box}(\cdot)$ is the box message packer that selectively packs individual detections to the box message conditioned on the budget, and $\Phi_{\rm point}(\cdot)$ is the point message packer that selectively incorporates raw observations into the point message, contingent upon individual detections and the communication budget.
$\Phi_{\rm fusion}(\cdot)$ is the fusion function that integrates the individual observation with the received collaborative messages to generate collaborative detections $\overline{\mathcal{O}}_i$.

Note that hybrid collaboration can be degraded to late and early collaboration. Specifically, when the box message packer in~\eqref{subeq:late} uses a heuristic all-or-nothing selection strategy, meaning it selects all detection results for transmission if bandwidth permits, otherwise it selects none, and bypasses equations~\eqref{subeq:early} and~\eqref{subeq:message}, hybrid collaboration is reduced to late collaboration. Similarly, hybrid collaboration shifts to early collaboration if the point message packer in~\eqref{subeq:early} applies the all-or-nothing strategy, and~\eqref{subeq:late}~\eqref{subeq:message} are overlooked.

Hybrid collaboration presents two distinctive benefits. First, hybrid collaboration takes advantage of multiple message types. The hybrid message packers dynamically select the most appropriate messages based on the available communication bandwidth, ensuring flexibility in response to varying communication constraints. Second, hybrid collaboration employs standardized box and point representation which are independent of specific detection models, promoting adaptability across diverse agent setups.


Specifically, \texttt{HyComm} includes: single detection with confidence and uncertainty estimation, achieving Equation~\eqref{subeq:detection} in Sec.~\ref{subsec:4a}; confidence-based box message packer, achieving Equation~\eqref{subeq:late} in Sec.~\ref{subsec:4b}; uncertainty-based observation message packer, achieving Equation~\eqref{subeq:early} in Sec.~\ref{subsec:4c}, and fusion module, achieving Equation~\eqref{subeq:fusion} in Sec.~\ref{subsec:4d}.

\vspace{-4mm}
\subsection{Single detection with confidence and uncertainty estimation}\label{subsec:4a}
\vspace{-1mm}
The single-agent detection with confidence and uncertainty estimation targets to generate object bounding boxes and estimate the associated confidence and uncertainty for each. They are used to guide the hybrid message packing. The intuition is that: i) the confidence score reflects the perceptually critical level, aiding in the prioritization of box messages; and ii) the uncertainty score reflects the precise level of box messages, directing the choice of additional observation messages that best complement the inaccurate box message. 

More specifically, we implement the single-agent 3D object detector $\Phi_{\rm detect}(\cdot)$ with the off-the-shelf designs, such as PointPillar\cite{PointPillar}, CenterPoint\cite{yin2021center}, to produce the estimated object bounding box. NMS is applied to achieve $K$ sparse detections $\mathcal{O}_i\in\mathbb{R}^{K\times 10}$ for the $i$th agent.
We parameterize each bounding box with confidence and uncertainty as $\mathbf{o}=(x,y,z,l,w,h,\theta,c,u_x,u_y)$, including the estimated 3D center positions, the length, width, height, yaw angle, the detection confidence score, and the uncertainty of the position of the detected objects in both the $x$ and the $y$ coordination. It is represented with position variance, this is, $u_x=\sigma_x^2,u_y=\sigma_y^2$. We denote the box regression, the confidence, and the uncertainty of all the sparse detections as $\mathbf{D}_i\in\mathbb{R}^{K\times 7}$, $\mathbf{C}_i\in[0,1]^{K}$ and $\mathbf{U}_i\in\mathbb{R}^{K\times 2}$. 
The estimated confidence and uncertainty will be used in the message packers to select the important boxes and points to be involved in the hybrid message. 


Note that: i) we represent the confidence score with the object detection confidence, which indicates the possibility of the existence of objects. During the collaboration, boxes containing objects are more critical and could help recover the miss-detected objects in a collaborator's limited view; and ii) we represent the uncertainty score with the object center regression variance, which indicates the potential displacement range of a box. During the collaboration, points contained within less precise boxes should be prioritized for transmission to supplement the limitations of the box message with additional perceptual information.

Three loss functions are jointly used to supervise the detection box learning, the total loss is 
$\label{eq:loss}
    L=L_{\rm reg}+L_{\rm cls}+L_{\rm unc},
$
where $L_{\rm reg}$ is the smooth-L1 loss for the box regression, $L_{\rm cls}$ is the cross-entropy loss for the object classification, serving for the confidence estimation, and $L_{\rm unc}$ is the Dirac function \cite{he2019bounding,lu2022robust} for the uncertainty estimation.

After the single-agent detection, considering the available communication bandwidth, agents pack and exchange hybrid messages to offer complementary information benefiting their partners. 
As illustrated in Equation~\eqref{subeq:message}, the hybrid message includes two types: box message and point message. 
The following two subsections introduce the implementation of box message packer and point message packer.

\vspace{-5mm}
\subsection{Confidence-based box message packer}\label{subsec:4b}
\vspace{-1mm}

When the communication budget is highly constrained, the transmission of efficient box messages takes precedence. The confidence-based box message packer achieves Equation~\eqref{subeq:late} with a novel box selection strategy.
It leverages the detection confidence score to reflect the perceptual significance and based on which to select the most informative boxes to the box messages within the communication budget. Intuitively, those boxes with lower confidence scores may be noisy predictions, such as false positives or backgrounds. Those boxes with higher confidence will provide more critical information to help the partners and should have a higher priority selected.

Following this spirit, we consider a proxy-constrained problem as follows,
\begin{equation}
\vspace{-1mm}
\underset{\mathbf{S}_i^{(o)}}{\max}~\sum\mathbf{S}_i^{(o)}\odot \mathbf{C}_i ~s.t.~ |\mathbf{S}_i^{(o)}|\leq b^{(o)},\mathbf{S}_i^{(o)}\in\{0,1\}^{K},\label{subeq:obj_select}
\vspace{-1mm}
\end{equation}
where $\mathbf{C}_i$ is the confidence score of the sparse detections, and $\mathbf{S}_i^{(o)}$ is the binary selection matrix that represents whether each of the $K$ detections is selected or not, where $1$ denotes selected, and $0$ elsewhere. Note that even though this optimization problem has hard constraints and non-differentialability of binary variables, it has an analytical solution that naturally satisfies all the constraints. The proof can be found in the appendix. This solution is obtained by selecting those boxes whose corresponding elements in $\mathbf{S}_i$ rank top-$b$. The detailed steps of \textbf{box selection} are: i) arrange the elements in the input in descending order; ii) given the communication budget constrain, decide the total number ($b^{(o)}={B}/{7}$) of available boxes; iii) set the spatial regions of $\mathbf{S}_i$, where elements rank in top-$b$ as the $1$ and $0$ verses. 


Based on the selection matrix, the selection function $\Phi_{\rm select}(\cdot)$ pack the selected boxes into the box messages by
\begin{equation}
\vspace{-1mm}
\mathcal{M}_i^{(o)}=\Phi_{\rm select}(\mathbf{S}_i^{(o)},\mathcal{O}_i)\in\mathbb{R}^{b^{(o)}\times 7}.
\end{equation}
The box message can not only complement the individual detections in the fusion module but also serve as key landmarks for pose error correction~\cite{lu2022robust}, promoting robustness.
Note that incorporating box-level selection in the box message packer is a novel design, departing from the conventional box message in late collaboration which relies on a heuristic all-or-nothing strategy, effective only when the budget permits transmitting all detections. This new design widens the application scope of collaborative perception by enabling adaptability in highly constrained communication conditions.

\vspace{-4mm}
\subsection{Uncertainty-based point message packer}\label{subsec:4c}


As the communication budget increases, additional informative point messages are incrementally incorporated to mitigate the limitations of the box message. For example, when there are sparse points on the object or when the object is partially obscured, the box message might be unreliable. In such cases, transmitting supplementary observation data is essential, allowing collaborative detection to outperform individual detection.

To select informative point messages, the uncertainty-based point message packer employs a novel point selection strategy, implementing Equation~\eqref{subeq:early}. It uses the uncertainty score to assess point significance, prioritizing the most important point data to complement box messages. The intuition is that uncertainty reflects potential displacement in box regression, where higher uncertainty indicates greater inaccuracy in the predicted box, requiring more raw points to convey comprehensive perceptual information. Thus, prioritizing points for more uncertain instances brings more beneficial perceptual information, optimizing the communication budget usage.

\begin{figure*}[!t]
    \includegraphics[width=1.0\linewidth]{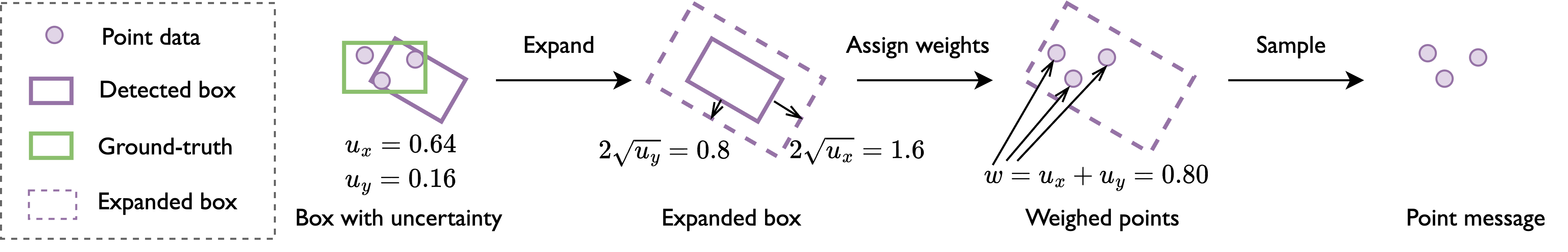}
    \vspace{-7mm}
    \caption{Overview of uncertainty-based point message packer. The point message packer processes inputs comprising the single-agent detections with uncertainty and the raw point clouds to generate the selected points for the point message. Initially, it expands the detected areas based on their associated uncertainty, creating a broader range of point candidates. Subsequently, these candidates are allocated weights that correspond to their level of uncertainty. Finally, a weighted sampling of these candidates is conducted to produce the point message.}
    \label{Fig:pointpacker}
    \vspace{-7mm}
\end{figure*}

Following this spirit, we achieve the uncertainty-based point message packer, see Fig.~\ref{Fig:pointpacker}. It involves three steps: i) expanding each inaccurate box based on its uncertainty to enhance coverage of raw points in regions where plausible instances may be located; ii) using uncertainty scores as sampling weights to select more raw points for uncertain instances; and iii) applying weighted sampling to pack selected points into point messages. Specifically, it takes detections $\mathbf{D}_i$ with uncertainty $\mathbf{U}_i$ as input and outputs point message $\mathcal{M}_i^{(x)}$ as
\begin{subequations}
\vspace{-2mm}
\begin{align}
    \mathbf{E}_i&=\Phi_{\rm expand}(\mathbf{D}_i,\mathbf{U}_i)\in\mathbb{R}^{D},\label{subeq:expand}\\
    \mathbf{W}_i&=\Phi_{\rm weight}(\mathcal{X}_i, \mathbf{E}_i,\mathbf{U}_i,\delta)\in\mathbb{R}^{P},\label{subeq:reweighting}\\
    \mathcal{M}_i^{(x)}&=\Phi_{\rm sample}(\mathcal{X}_i,\mathbf{W}_i,b^{(x)})\in\mathbb{R}^{b^{(x)}\times 4}. \label{subeq:sampling}
\end{align}  
\end{subequations}
In Equation\eqref{subeq:expand}, the expanding function $\Phi_{\rm expand}$ generates the expanded detections $\mathbf{E}_i$ to account for the position estimation uncertainty, this is, the $k$th enlarged box is $\mathbf{E}_i[k]=(x,y,z,l+2\sqrt{u_x},w+2\sqrt{u_y},h,\theta,c)$, where the parameters are from $\mathbf{D}_i[k]$. Expanding the less accurate box to a larger region allows for involving more raw observations to help locate the instance.
In Equation\eqref{subeq:reweighting}, the weight function $\Phi_{\rm weight}$ assigns a weight to each observed point cloud in $\mathcal{X}_i\in\mathbb{R}^{P\times 4}$, resulting in the weight matrix $\mathbf{W}_i$. Points situated within the expanded detections $\mathbf{E}_i$ are assigned with a weight corresponding to the instance's uncertainty score, this is, $u_x+u_y$. Others are assigned with a minimal default weight $\delta$. This assignment emphasizes the notion that these points may deliver supplemental information for instances identified with less accuracy.
In Equation\eqref{subeq:sampling}, the sampling function $\Phi_{\rm sample}$ employs a weighted sampling algorithm to sample $b^{(x)}$ points within the communication budget, which is what remains post the transmission of box messages, this is, $b^{(x)}={(B-7K)}/{4}$. 
By doing so, we can procure informative supplementary point messages to compensate for box messages. Then, agents exchange these selected hybrid messages.

\noindent\textbf{Novelty over previous collaboration strategies.} 
Compared to existing hybrid collaboration strategy~\cite{ArnoldDT:22}, which relies on manually setting the sensor radius as a hyper-parameter to heuristically switch between early and late collaboration and transmit all the data in each collaboration mode, our uncertainty-guided hybrid collaboration has two key innovations: i) it leverages box uncertainty as a guiding mechanism to adaptively switch between late and early collaboration, and ii) it prioritizes perceptually critical data parts in each collaboration mode. 
By adaptively supplementing less accurate boxes with precise raw point data, our method effectively mitigates the potential errors in late collaboration with early collaboration, ensuring robust performance. 
By selectively prioritizing highly confident box messages and points within uncertain boxes, our method allocates the communication budget to the most critical data parts, promoting communication efficiency.
As a result, our method delivers a superior performance-communication trade-off and significantly reduces potential error risks associated with combining different collaboration types.

\vspace{-5mm}
\subsection{Fusion}\label{subsec:4d}
\vspace{-1mm}
The fusion module enhances the detections by aggregating the received hybrid messages. Point messages are fused with the individual observations to produce the upgraded detections, which is the standard early fusion. The upgraded detections are then fused with the box messages to generate the final collaborative detections, which is the standard late fusion. Specifically, for $i$-th agent, given its raw point clouds $\mathcal{X}_i$, the received point message $\mathcal{M}_j^{(x)}$ and box message $\mathcal{M}_j^{(o)}$ from agent $j$, the process is given by
\begin{subequations}
\vspace{-2mm}
\begin{align}
\widetilde{\mathcal{O}}_i&=\Phi_{\rm detect}(\mathcal{X}_i \cup \{\mathcal{M}_j^{(x)}\}_{j\in\mathcal{N}_i}), \label{subeq:early_fusion} \\
\overline{\mathcal{O}}_i&=\Phi_{\rm nms}(\widetilde{\mathcal{O}}_i \cup \{\mathcal{M}_j^{(o)}\}_{j\in\mathcal{N}_i}), \label{subeq:late_fusion}
\end{align}
\end{subequations}
where $\cup$ denotes the union operation, $\widetilde{\mathcal{O}}_i$ is the updated detections after early fusion, this is, re-detect the merged point clouds.  $\overline{\mathcal{O}}_i$ is the final collaborative detections after late fusion, this is, merge the boxes and apply the NMS operation $\Phi_{\rm nms}(\cdot)$ to eliminate the redundant detections. 

\noindent\textbf{Advantages of our hybrid collaboration.}
\texttt{HyComm} allows collaborative perception to take advantage of the benefits of various message types, combining the compact advantage of box messages with the informative advantage of point messages. 
It offers three distinct advantages: i) allows for flexible adjustment of sampling rates, catering to a range of communication budgets from extremely limited to moderate and ample, consistently promoting communication efficiency, ii) the usage of a standard, uniform representation ensures independence from specific detection models, thus fostering broader applicability and generalizability, and iii) the included box messages can serve as landmarks, facilitating pose error correction and thereby bolstering robustness.

\vspace{-4mm}
\section{Experiments}
\label{sec:experiments}
\vspace{-1mm}

We validate \texttt{HyComm} in collaborative LiDAR-based 3D object detection tasks on the real-world benchmark DAIR-V2X and the simulation benchmark OPV2V.

\vspace{-4mm}
\subsection{Dataset and implementation details}
\vspace{-1mm}

\textbf{DAIR-V2X}~\cite{dair} is a \textbf{real-world} vehicle-to-vehicle and infrastructure collaborative perception dataset. The perception range is $x\in[-100\text{m}, 100\text{m}],y\in[-40\text{m}, 40\text{m}]$. \textbf{OPV2V}~\cite{OPV2V} is a vehicle-to-vehicle collaborative perception dataset. The perception range is $x\in[-140\text{m}, 140\text{m}],y\in[-40\text{m}, 40\text{m}]$. 
Our detector follows PointPillar~\cite{PointPillar} and CenterPoint~\cite{yin2021center}. 
For the \textbf{pose error setting}, we follow CoAlign~\cite{lu2022robust}, adding Gaussian noise $\mathcal{N}(0,0.3)$ on 2D centers $x,y$ and yaw angle $\theta$ to the accurate global poses.
For the \textbf{heterogeneous setting}, we equip the ego agent and collaborators with different model parameters while sharing the same PointPillar~\cite{PointPillar} structure.


HyComm is trained with optimizer Adam, using learning rate 2e-3. The communication volume follows the standard setting as~\cite{hu2022where2comm,disconet,HuCollaboration:CVPR23,OPV2V} that counts the message size by byte in log scale with base $2$. We do not consider any extra data/feature/model compression to compare communication results straightforwardly and fairly. Mathematically, the communication volume for each box is calculated as $\text{log}_2\left(7 \times 32 / 8\right)$ bytes, and for each selected point, it is $\text{log}_2\left(4 \times 32 / 8\right)$ bytes. Here, the numbers 7 and 4 represent the quantity of floating-point numbers required to describe each box and point, respectively. The factor of $32$ is included because each number is represented using the float32 data type, and the division by $8$ converts bits to bytes.

\textbf{Adaptation to varying communication bandwidth budgets.} In realistic scenarios, bandwidth is often constrained and dynamic but can be estimated using available technologies. Given the bandwidth budget, HyComm initially prioritizes boxes with higher confidence. If bandwidth allows for more than just boxes, the remaining capacity calculates the total number of points that can be transmitted. Points are then weighted and sampled based on their uncertainty scores, with higher uncertainty points given priority. Finally, the selected boxes and sampled points are packed into messages for transmission to collaborators.


\begin{figure*}[!t]
  \centering
    \includegraphics[width=1.0\linewidth]{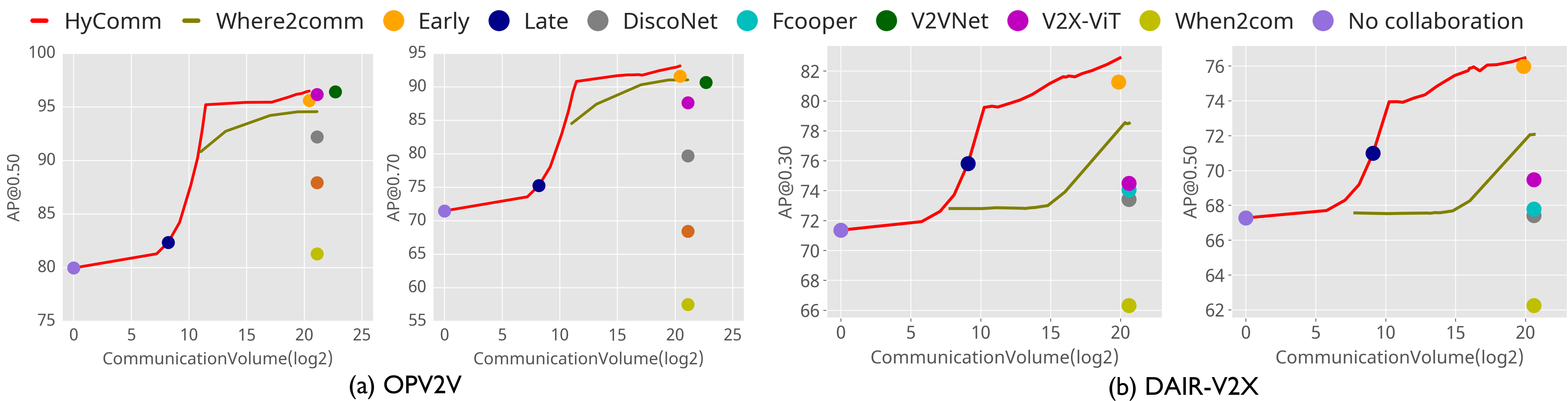}
    \vspace{-6mm}
  \caption{\texttt{HyComm} consistently achieves superior performance-bandwidth trade-off on both the simulated and real-world datasets, outperforming Where2comm with 2,006/1,317 $\times$ less communication cost on DAIR-V2X/OPV2V for AP50/70.}
  \vspace{-6mm}
  \label{Fig:SOTAs}
\end{figure*}

\vspace{-3mm}
\subsection{Quantitative results}

\textbf{Benchmark comparison.} Fig.~\ref{Fig:SOTAs} compares \texttt{HyComm} with the previous methods in terms of the trade-off between detection performance and communication bandwidth on both the real-world dataset, DAIR-V2X, and simulation dataset, OPV2V. The detection performance is measured with AP at IOU (intersection over union) 0.30, 0.50, and 0.70, and the communication cost is measured by byte in the log scale.
We consider no collaboration ($\mathcal{O}_i$), late collaboration, early collaboration, and intermediate collaboration, including Where2comm~\cite{hu2022where2comm}, Fcooper~\cite{Chen2019FcooperFB}, When2com~\cite{when2com}, V2VNet~\cite{v2vnet}, DiscoNet~\cite{disconet}, and V2X-ViT~\cite{xu2022v2x}. The red curve comes from a single~\texttt{HyComm} model evaluated at varying bandwidths. We see that the proposed \texttt{HyComm}: i) achieves a far-more superior perception-communication trade-off across the entire spectrum of communication bandwidth, and ii) achieves more than 2,006$\times$ and 1,317$\times$ lower communication volume and still outperforms previous SOTA Where2comm on both real-world and simulation scenarios.

\textbf{Adaptability across the whole communication spectrum.} In Fig.~\ref{Fig:SOTAs}, only the proposed \texttt{HyComm} and intermediate collaboration method ~\texttt{Where2comm} are depicted with curves while other existing methods are represented with a single point. This distinction highlights the adaptability of \texttt{HyComm} and \texttt{Where2comm} compared to the limited flexibility of other approaches. The reason for this adaptability is that \texttt{HyComm} and \texttt{Where2comm} incorporates a novel design that allows for message selection in communication at the box-level, point-level, and feature-level. In contrast, other collaborative methods rely on a heuristic all-or-nothing approach, which is only effective when bandwidth is sufficient to support the transmission of all detections, features, or point clouds. 

\begin{figure}[!t]
    \centering
    \vspace{1mm}
    \includegraphics[width=\linewidth]{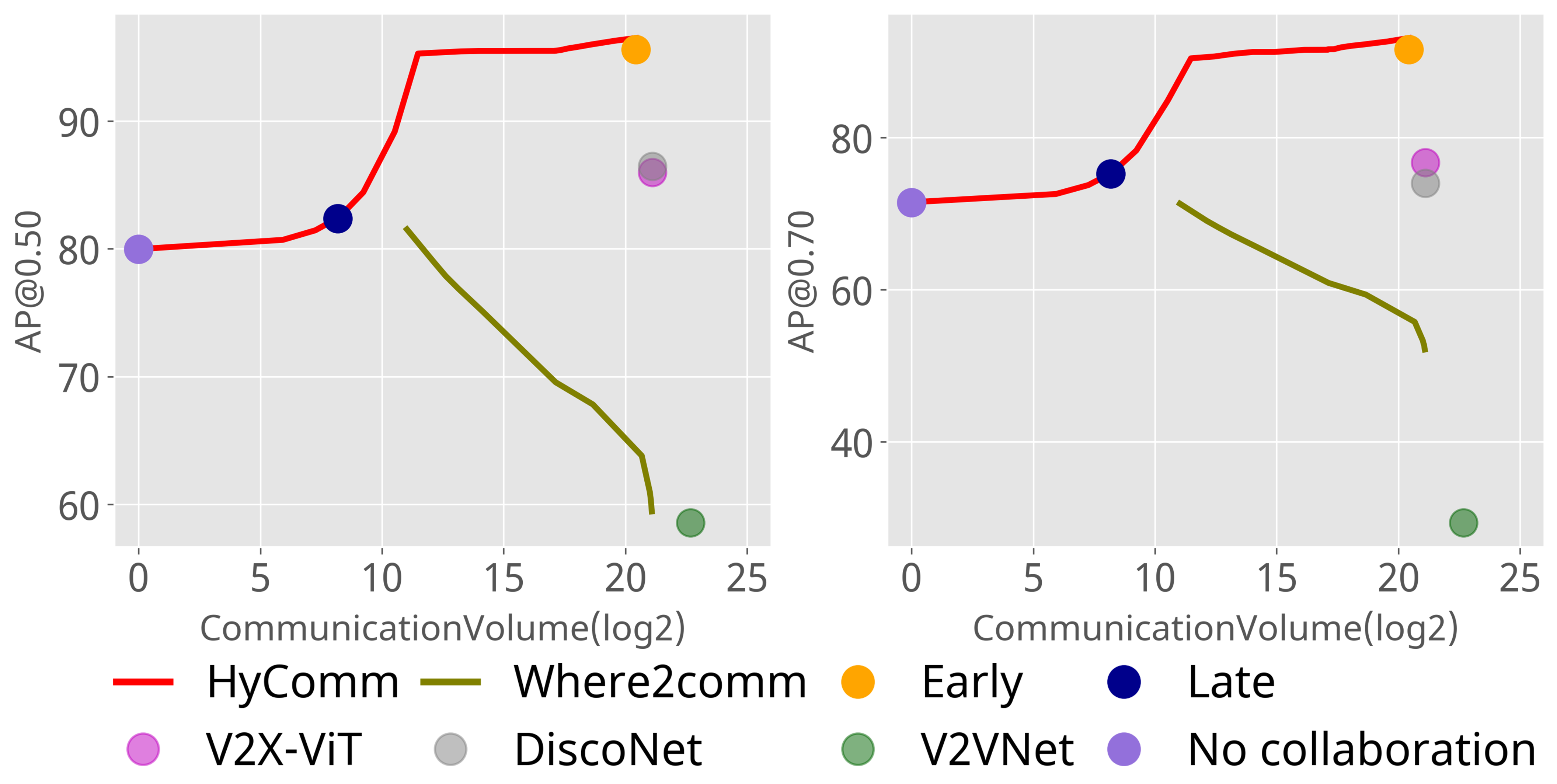}
    \vspace{-6mm}
  \caption{ Under the scenario that agents are equipped with different models, \texttt{HyComm} maintains a superior perception-bandwidth trade-off, while intermediate collaboration suffers from the misalignment between fused feature vectors, performing worse when more data are transmitted.}
  \vspace{-6mm}
  \label{Fig:Hete}
\end{figure}


\textbf{Generalizability under heterogeneous scenarios.} 
Fig.~\ref{Fig:Hete} compares \texttt{HyComm} with the previous methods under heterogeneous scenarios on OPV2V, where the ego agent and the collaborators are equipped with different models, using the same PointPillar~\cite{PointPillar} structure with different parameters. We see that: i) \texttt{HyComm} consistently maintains a superior perception-bandwidth trade-off, ii) intermediate collaboration under heterogeneous scenarios experiences significant degradation compared to performance under homogeneous scenarios, as depicted in Fig.~\ref{Fig:SOTAs}, and iii) as more communication information is received, intermediate collaboration performance degrades consistently, dropping below the level achieved by single detections alone. The reason is that hybrid box and point messages are standardized, naturally model agnostic representations. In contrast, features encoded by different models exhibit significant domain gaps, which intermediate collaboration cannot directly address. Messages with domain gaps introduce noisy and disruptive information, thus impacting overall performance.


\begin{figure*}[!t]
  \centering
    \includegraphics[width=1.0\linewidth]{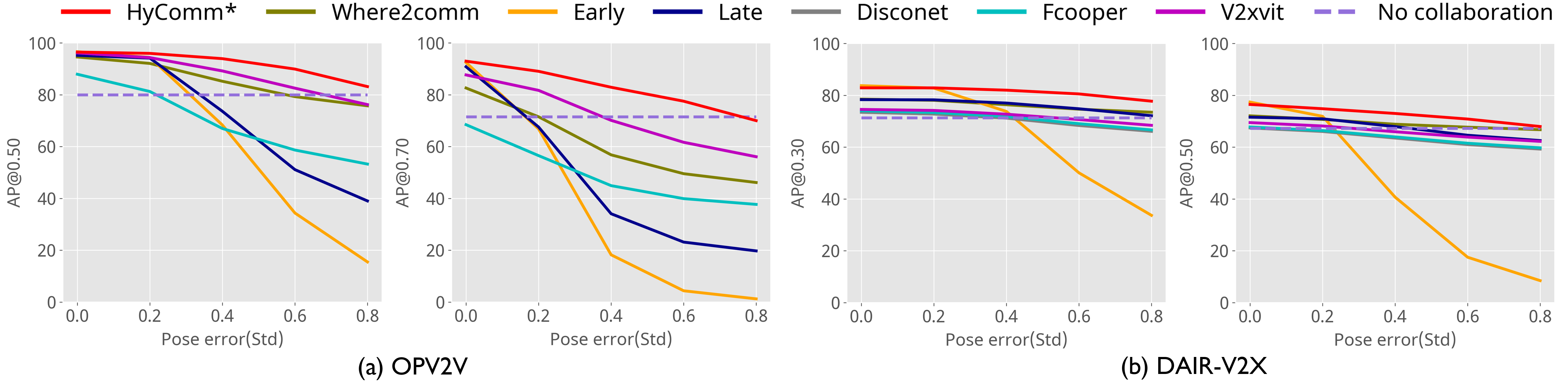}
    \vspace{-6mm}
  \caption{Robustness to pose error. \texttt{HyComm$^*$} with pose-robust design~\cite{lu2022robust} consistently outperforms previous SOTAs.}
  \vspace{-6mm}
  \label{Fig:PoseError}
  \vspace{-1mm}
\end{figure*}

\textbf{Robustness to pose error.} Fig.~\ref{Fig:PoseError} shows the detection performance in the presence of pose errors. The pose error setting follows~\cite{lu2022robust,v2vnet,xu2022v2x} using Gaussian noise with a mean of 0m and standard deviations ranging from 0m to 0.6m.
\texttt{HyComm$^*$} denotes our method aligned with the post-robust design~\cite{lu2022robust}, as the box messages can serve as landmarks for pose correction. 
We see that: i) \texttt{HyComm$^*$} is robust while other methods experience significant performance reductions with increasing levels of pose error; ii) \texttt{HyComm$^*$} consistently outperforms baselines under all imperfect conditions; and iii) \texttt{HyComm$^*$} consistently surpasses No Collaboration, whereas baselines fail when pose error exceeds 0.6m. 



\vspace{-4mm}
\subsection{Ablation studies}
\vspace{-1mm}


\begin{figure}[!t]
  \centering
    \includegraphics[width=0.46\linewidth]{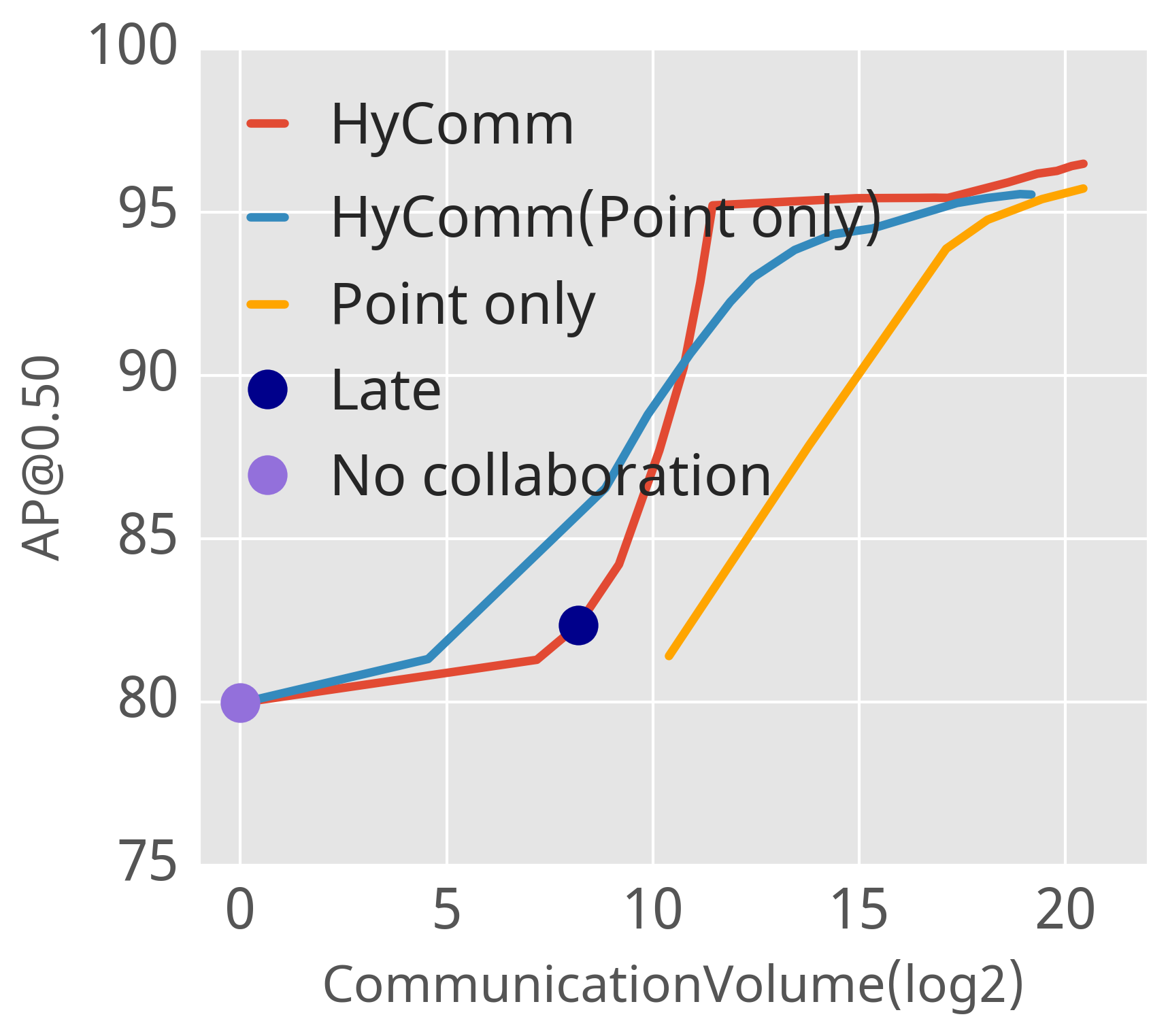}
    \includegraphics[width=0.45\linewidth]{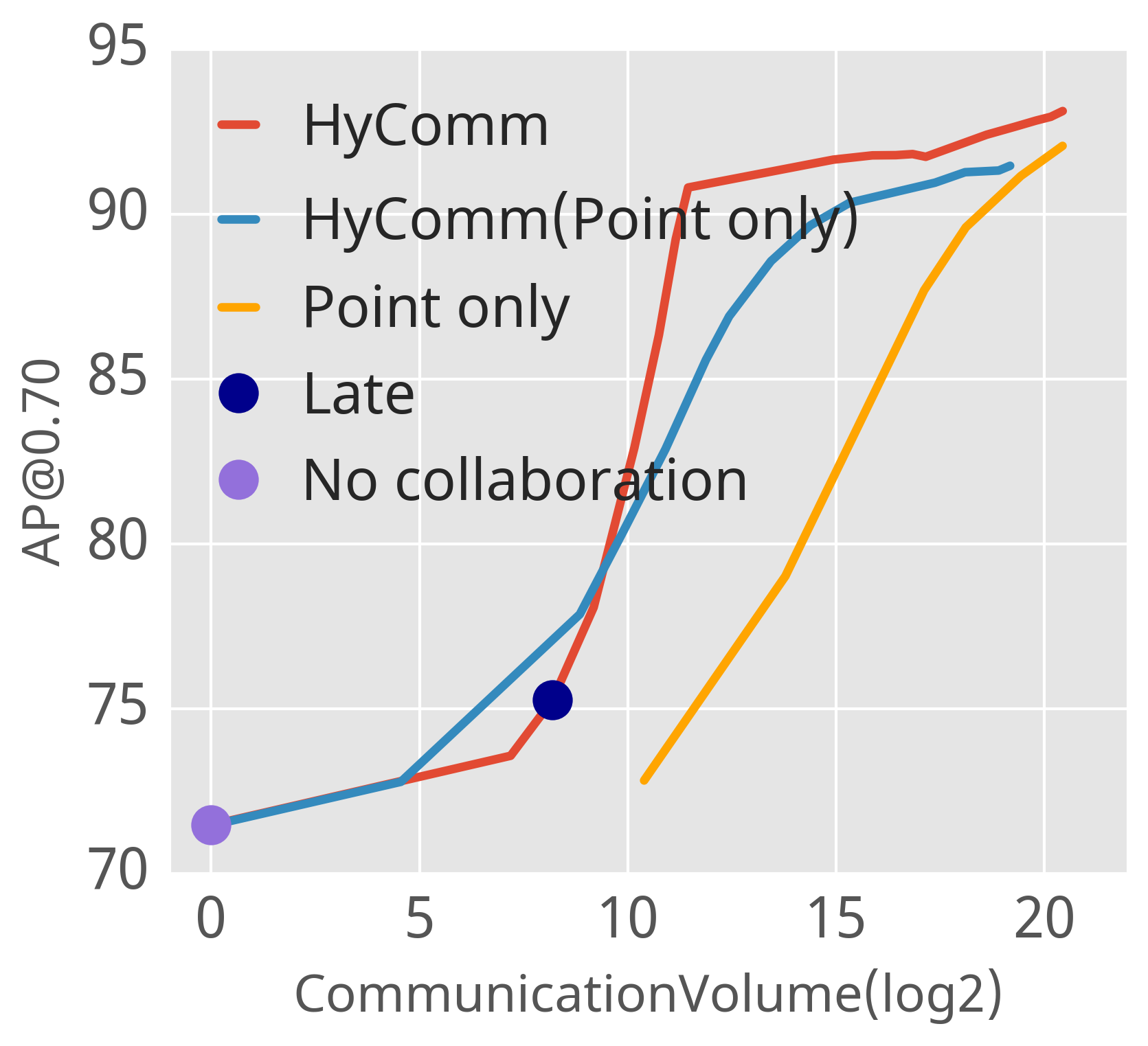}
  \vspace{-2mm}
  \caption{Ablation studies on hybrid collaboration and selection.}
  \vspace{-5mm}
  \label{Fig:hybrid}   
\end{figure}

\begin{figure}[!t]
  \centering
    \includegraphics[width=0.47\linewidth]{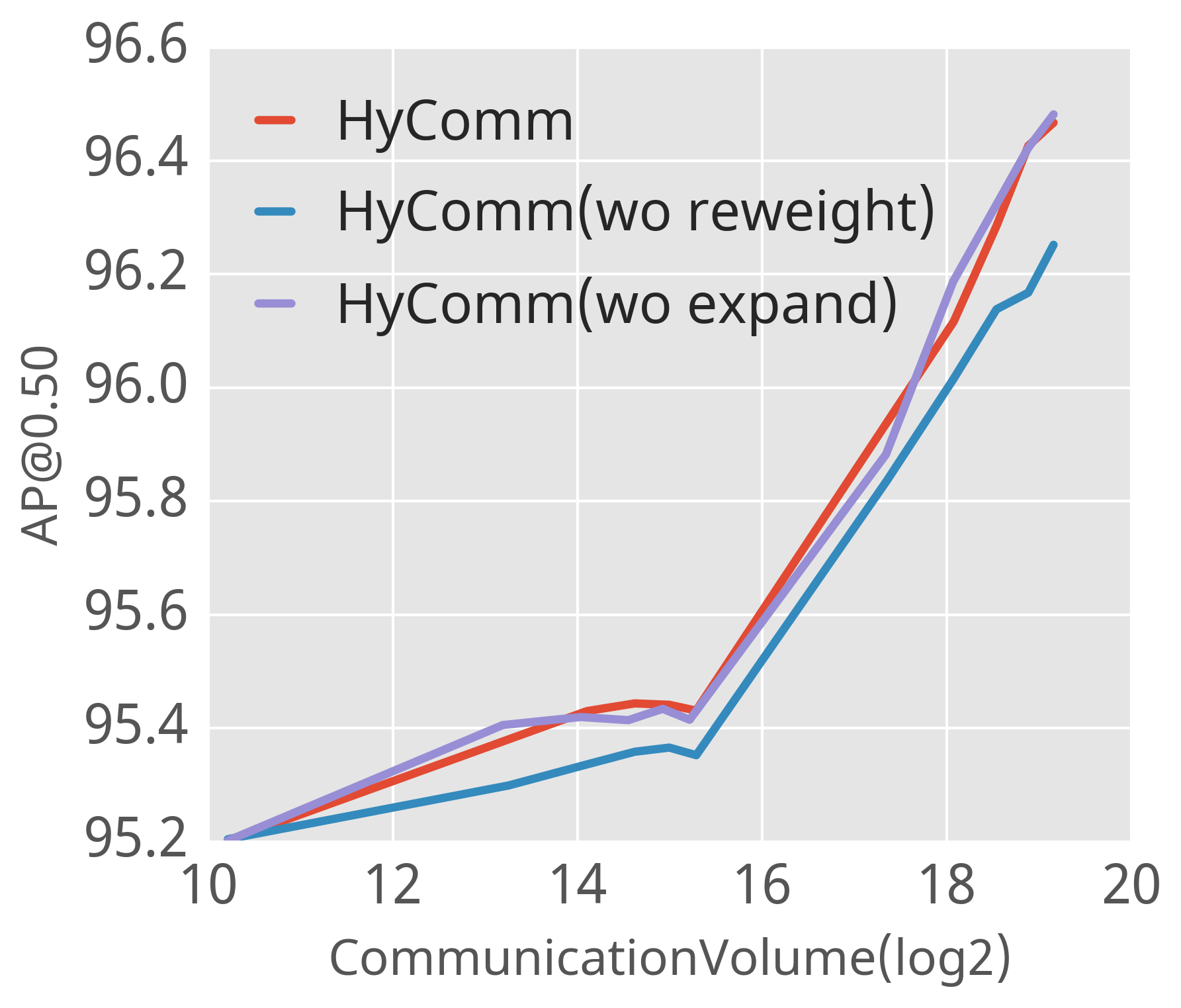}
    \includegraphics[width=0.47\linewidth]{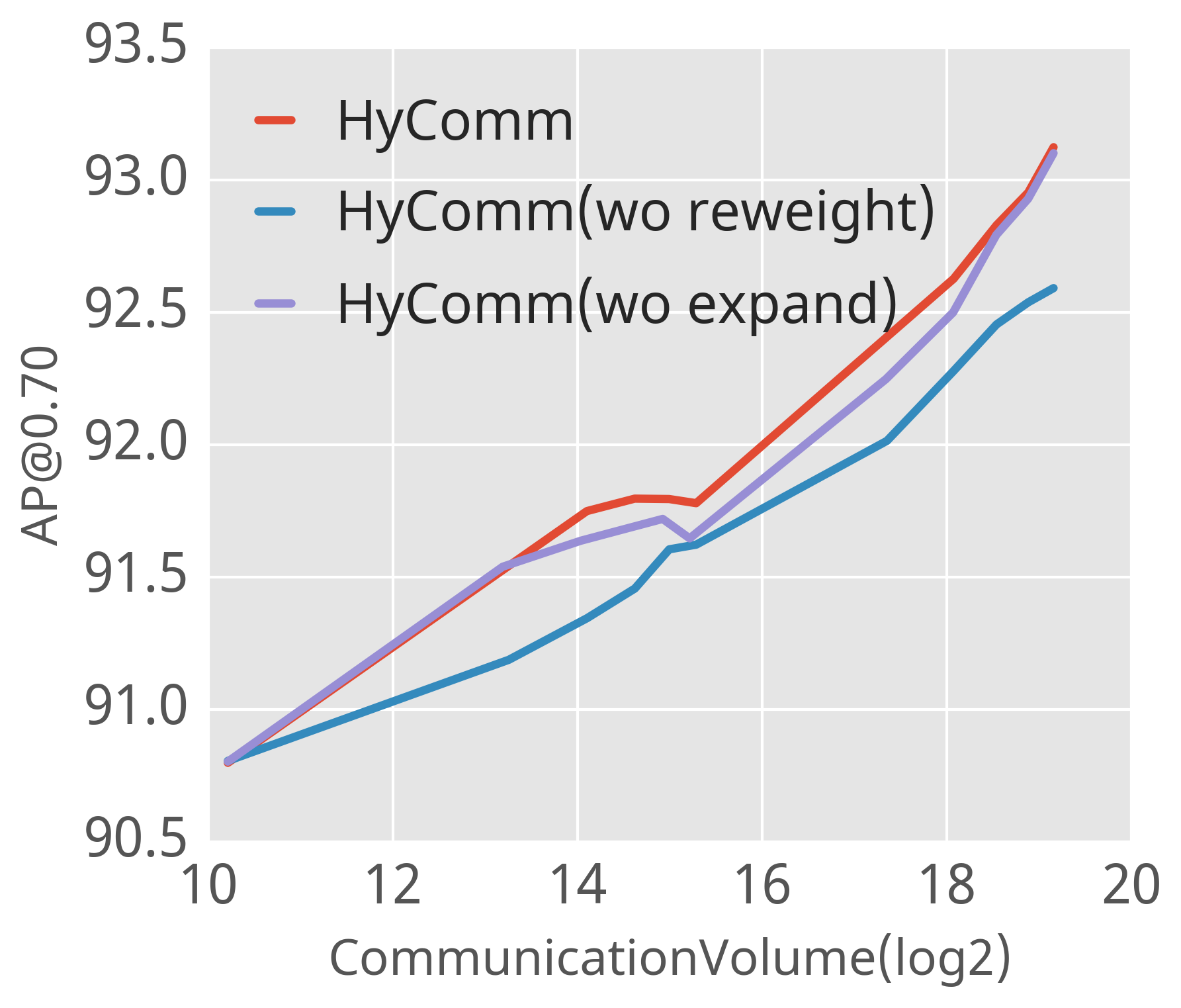}
  \vspace{-2mm}
  \caption{Ablation studies on uncertainty-based selection.}
  \vspace{-7mm}
  \label{Fig:uncertainty}
\end{figure}

\begin{figure}[!t]
  \centering
  \begin{subfigure}{1.0\linewidth}
    \includegraphics[width=0.46\linewidth]{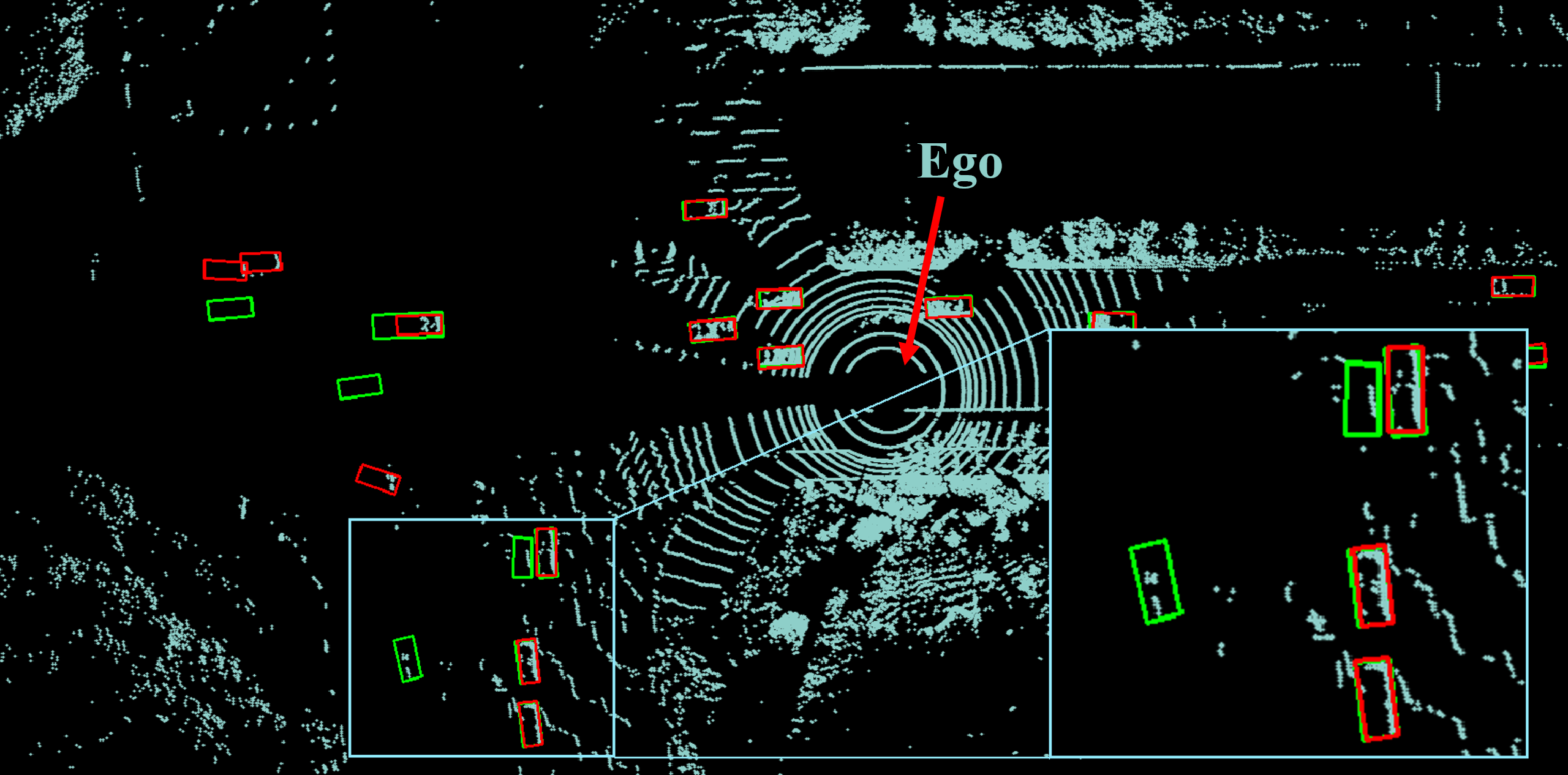}
    \includegraphics[width=0.49\linewidth]{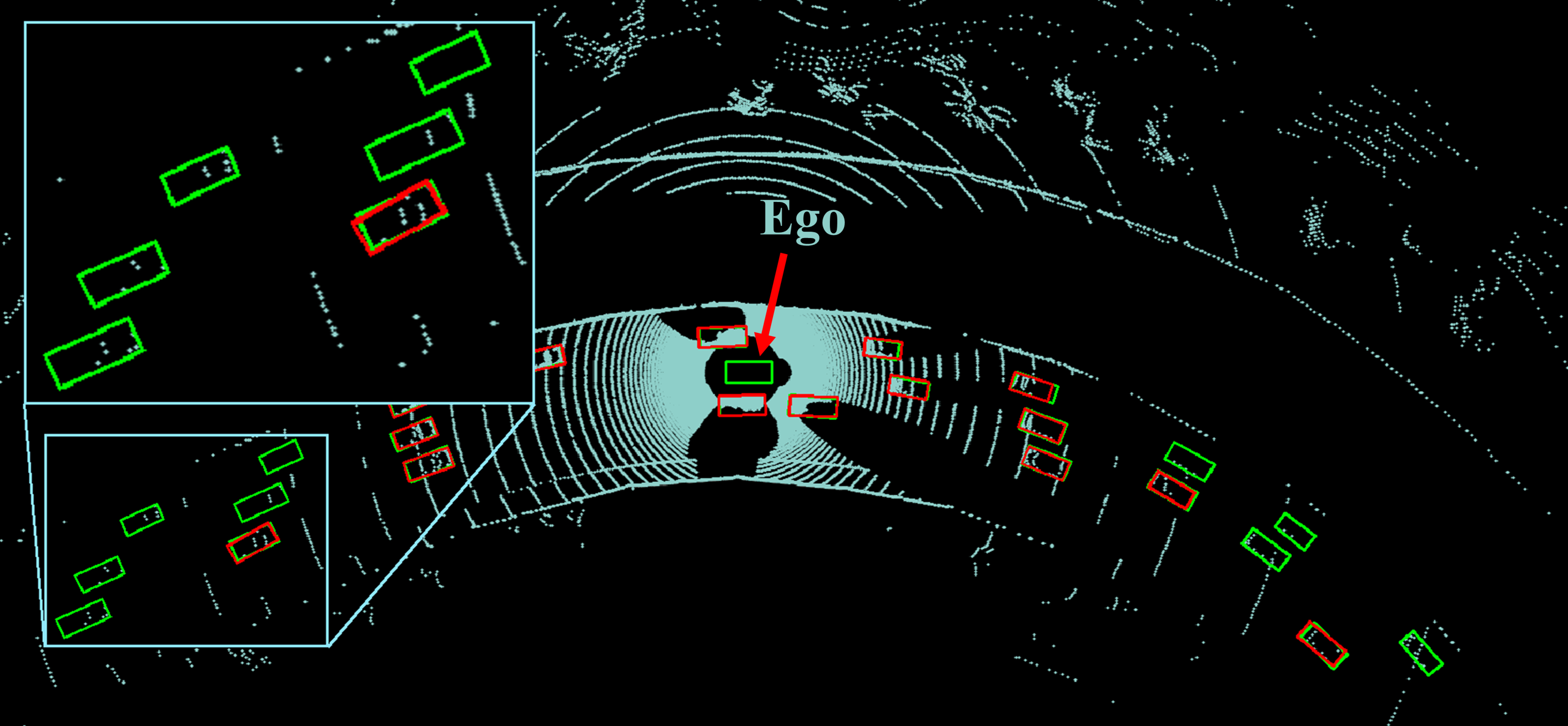}
    \vspace{-1mm}
    \caption{No Collaboration}
  \end{subfigure}
  \begin{subfigure}{1.0\linewidth}
    \includegraphics[width=0.46\linewidth]{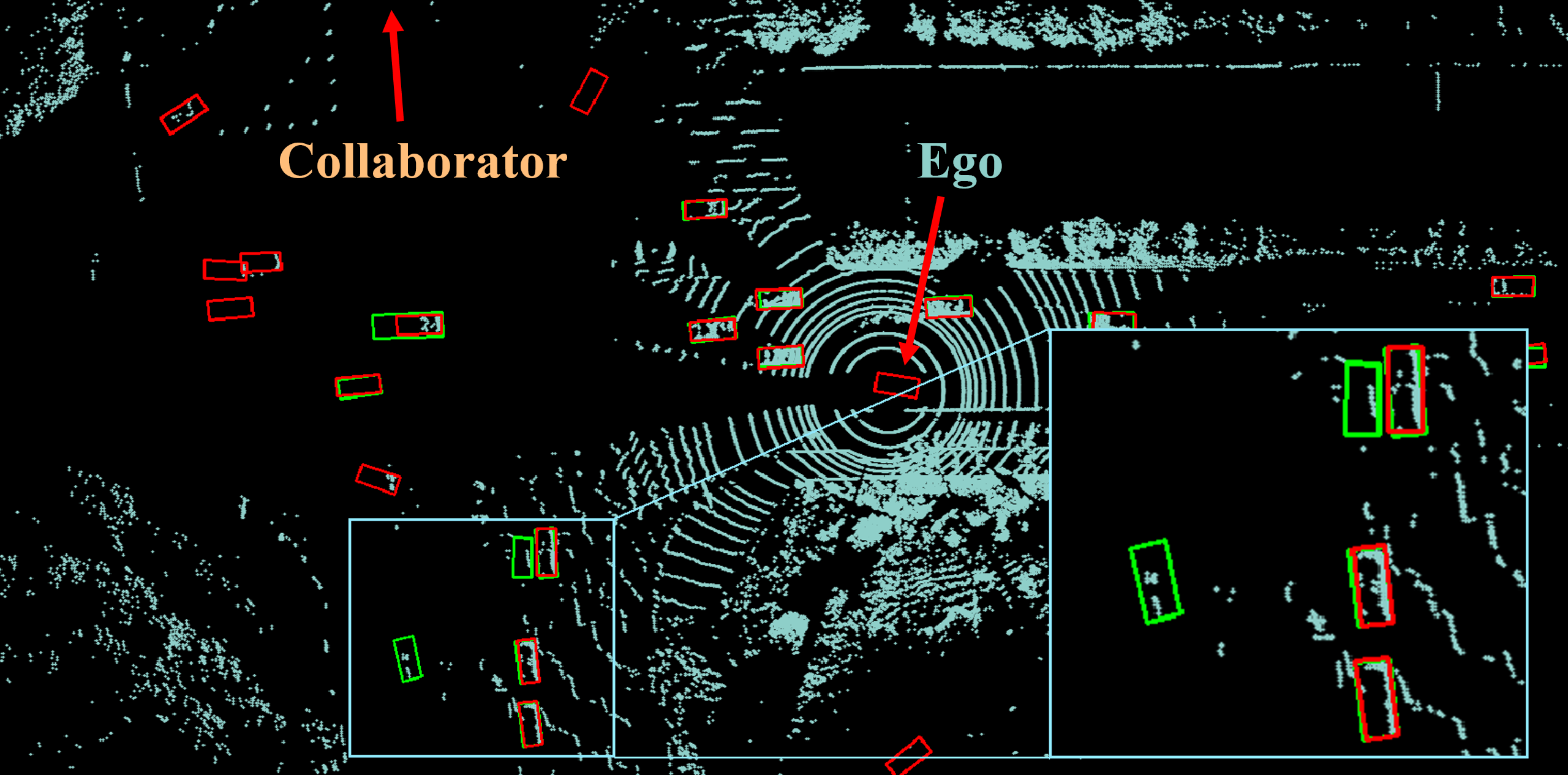}
    \includegraphics[width=0.49\linewidth]{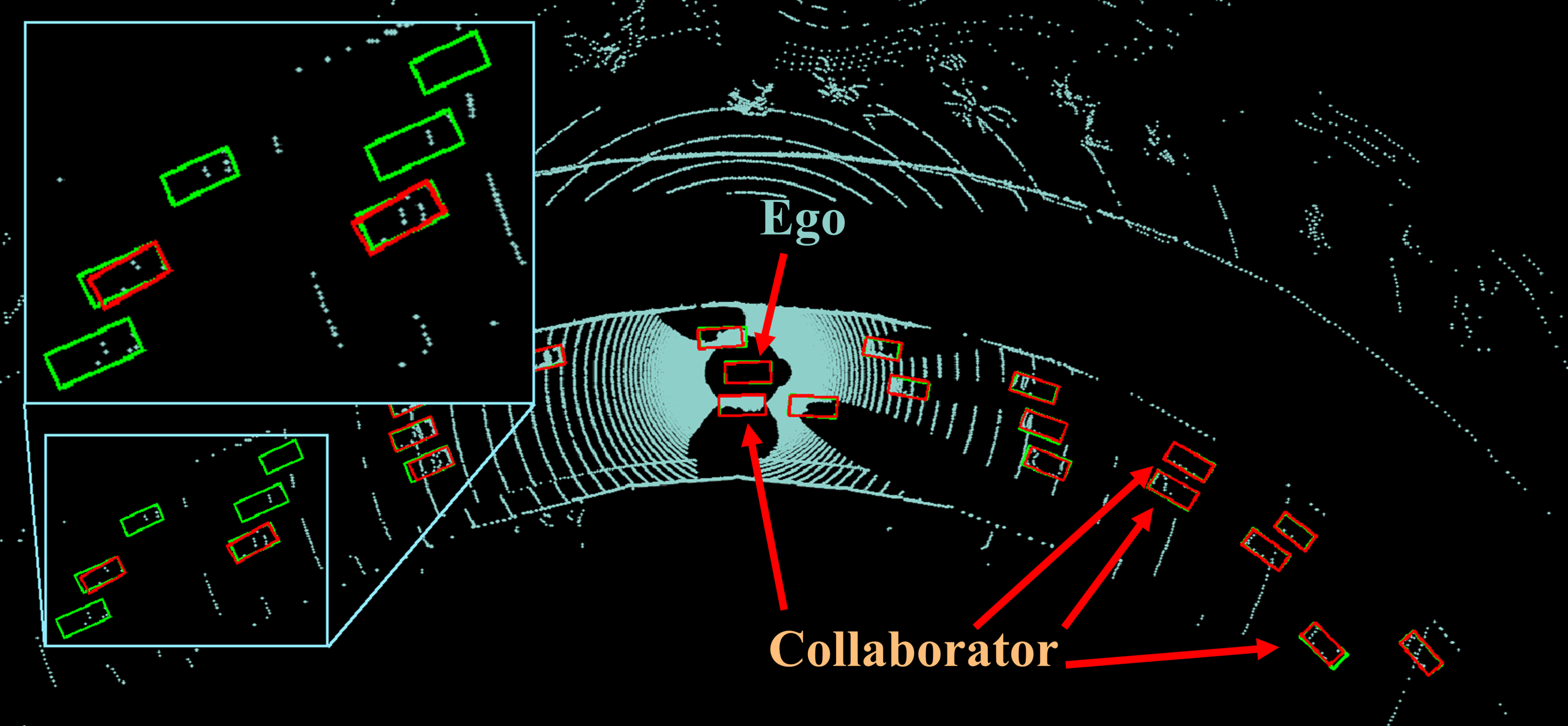}
    \vspace{-1mm}
    \caption{Late}
  \end{subfigure}
  \begin{subfigure}{1.0\linewidth}
    \includegraphics[width=0.46\linewidth]{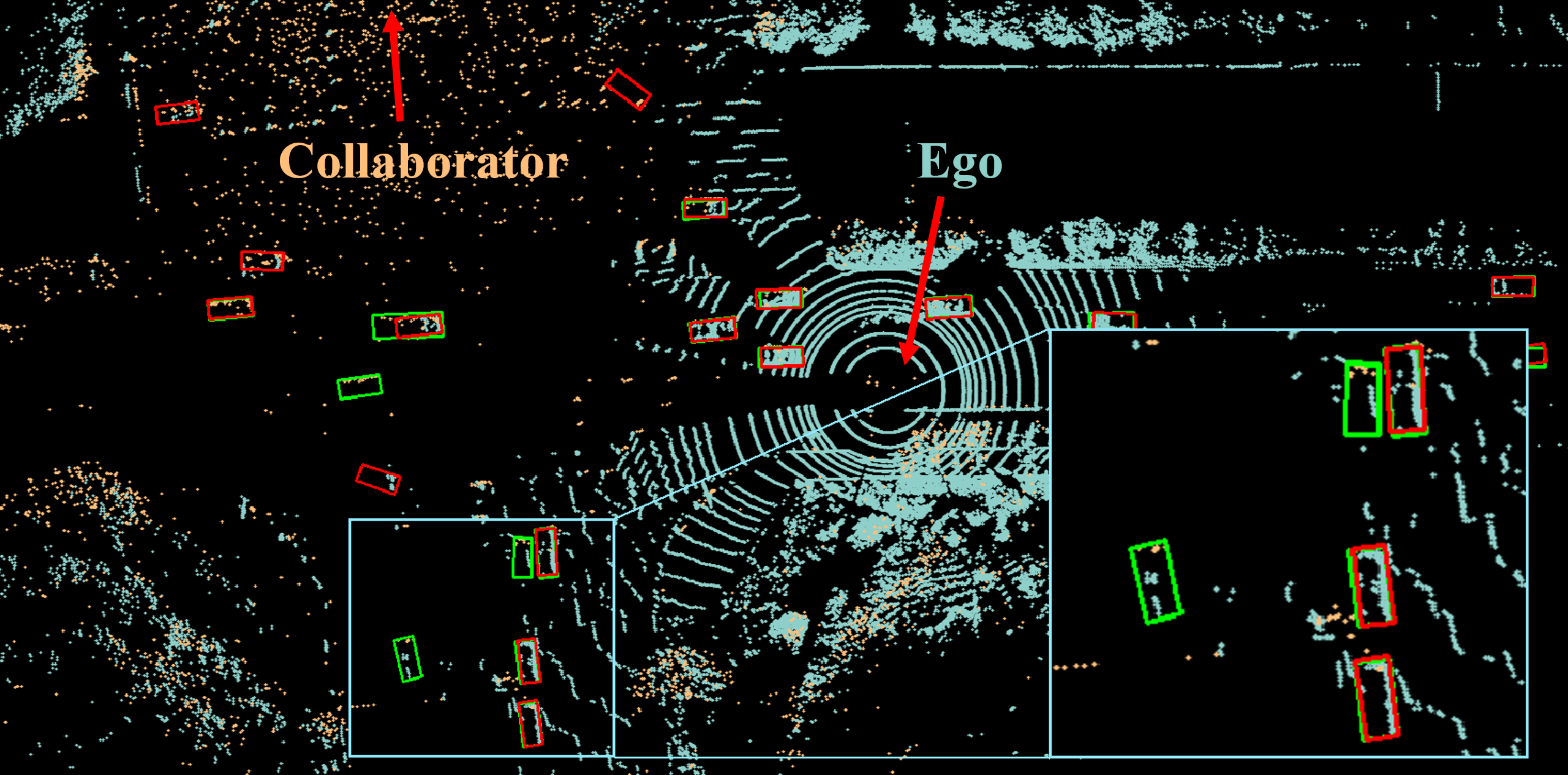}
    \includegraphics[width=0.49\linewidth]{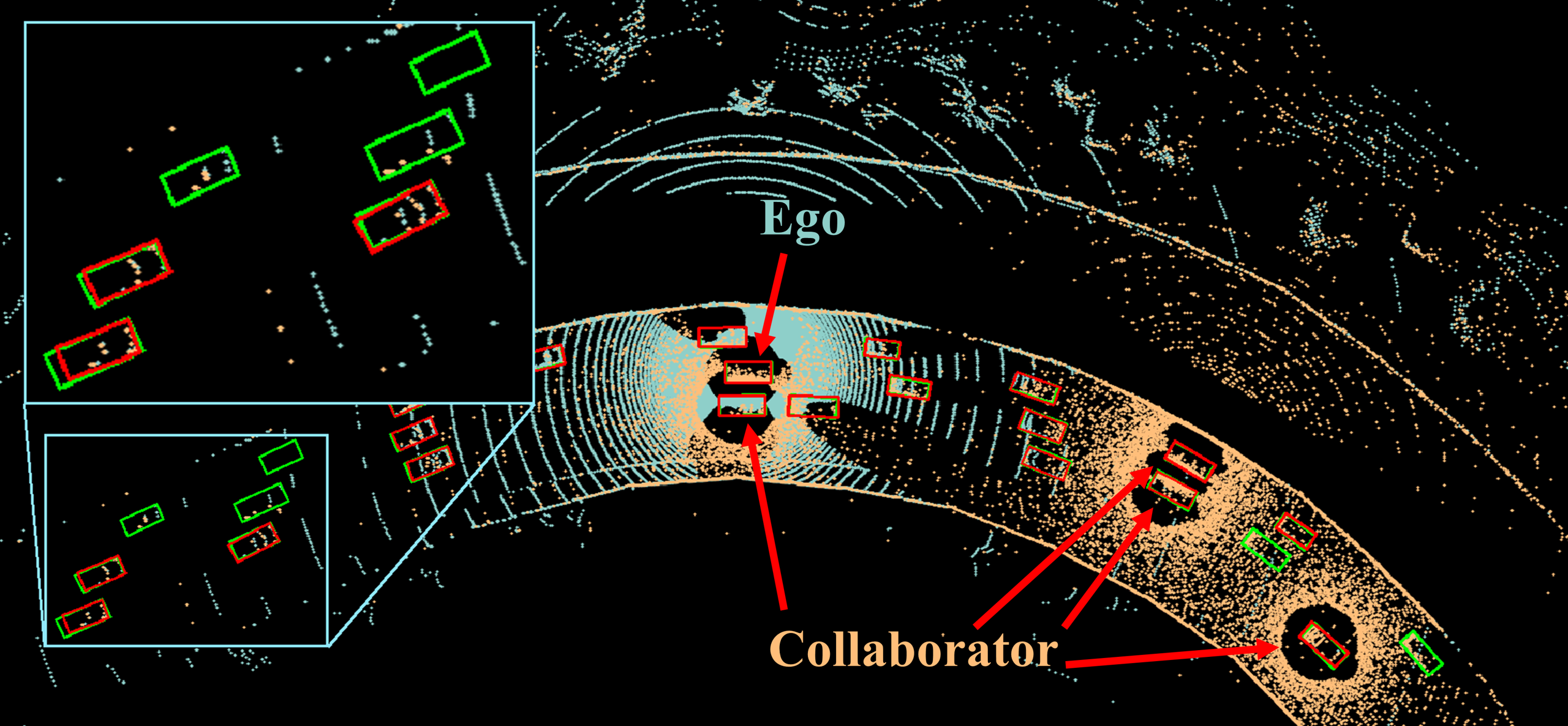}
    \vspace{-1mm}
    \caption{Early}
  \end{subfigure}
  \begin{subfigure}{1.0\linewidth}
    \includegraphics[width=0.46\linewidth]{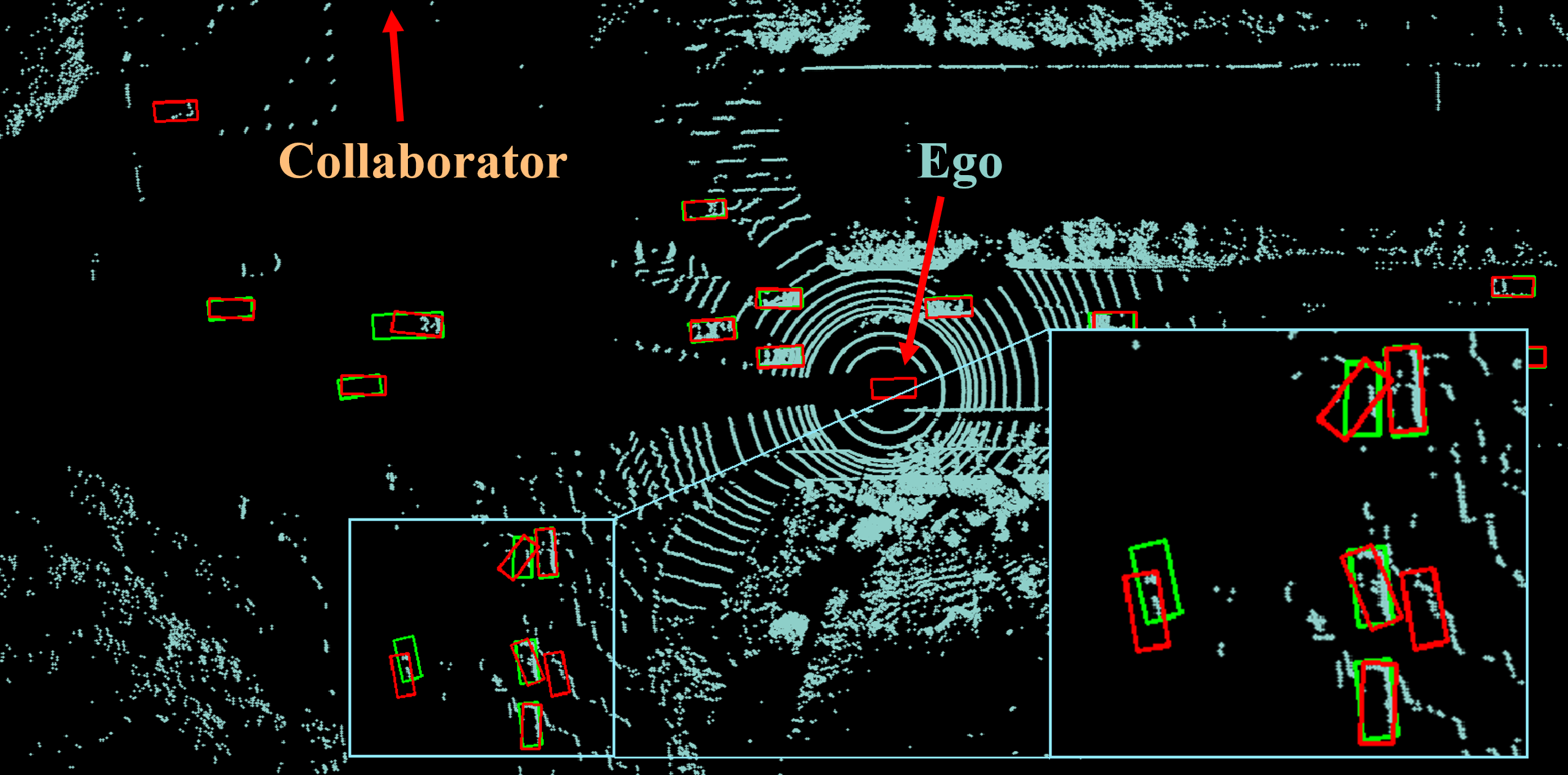}
    \includegraphics[width=0.49\linewidth]{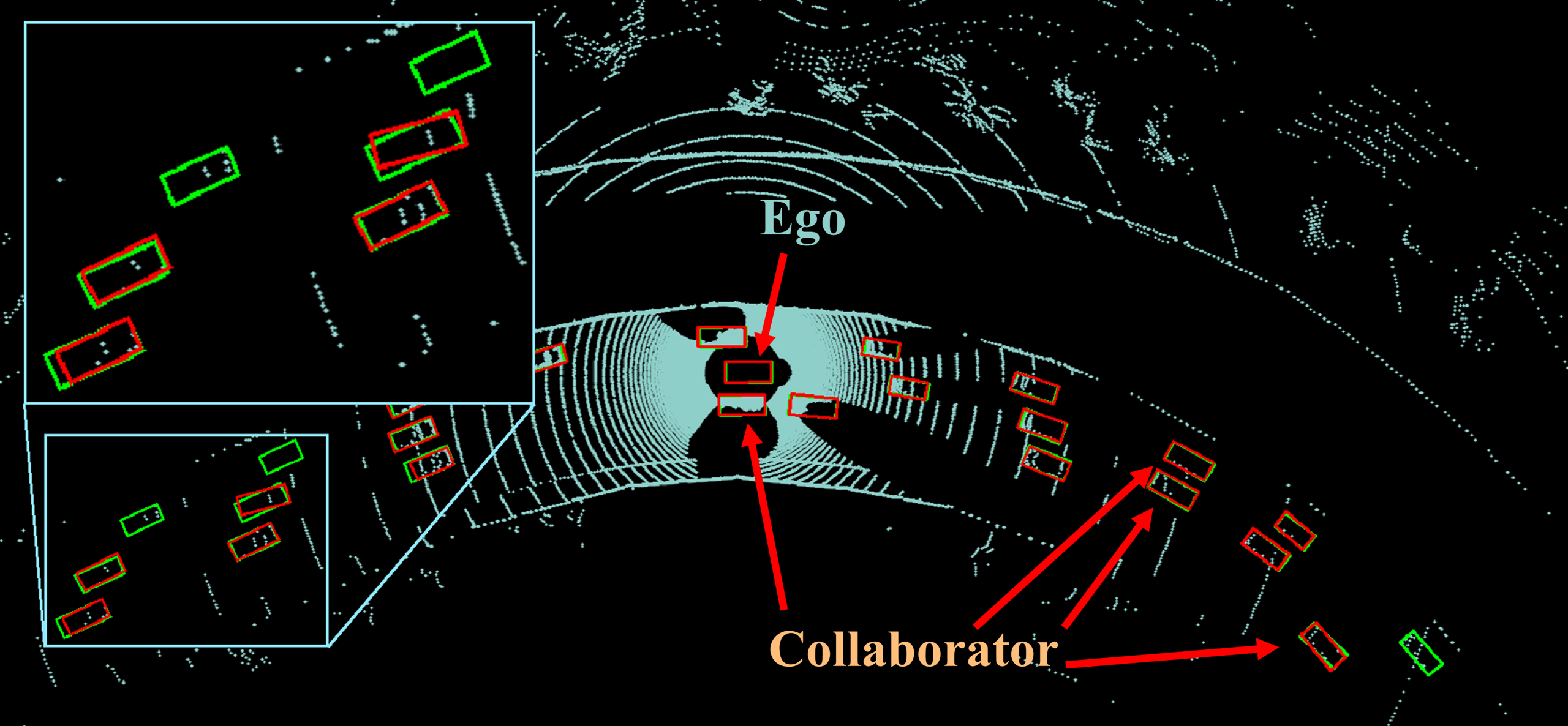}
    \vspace{-1mm}
    \caption{Where2comm (Intermediate)}
  \end{subfigure}
 \begin{subfigure}{1.0\linewidth}
    \includegraphics[width=0.46\linewidth]{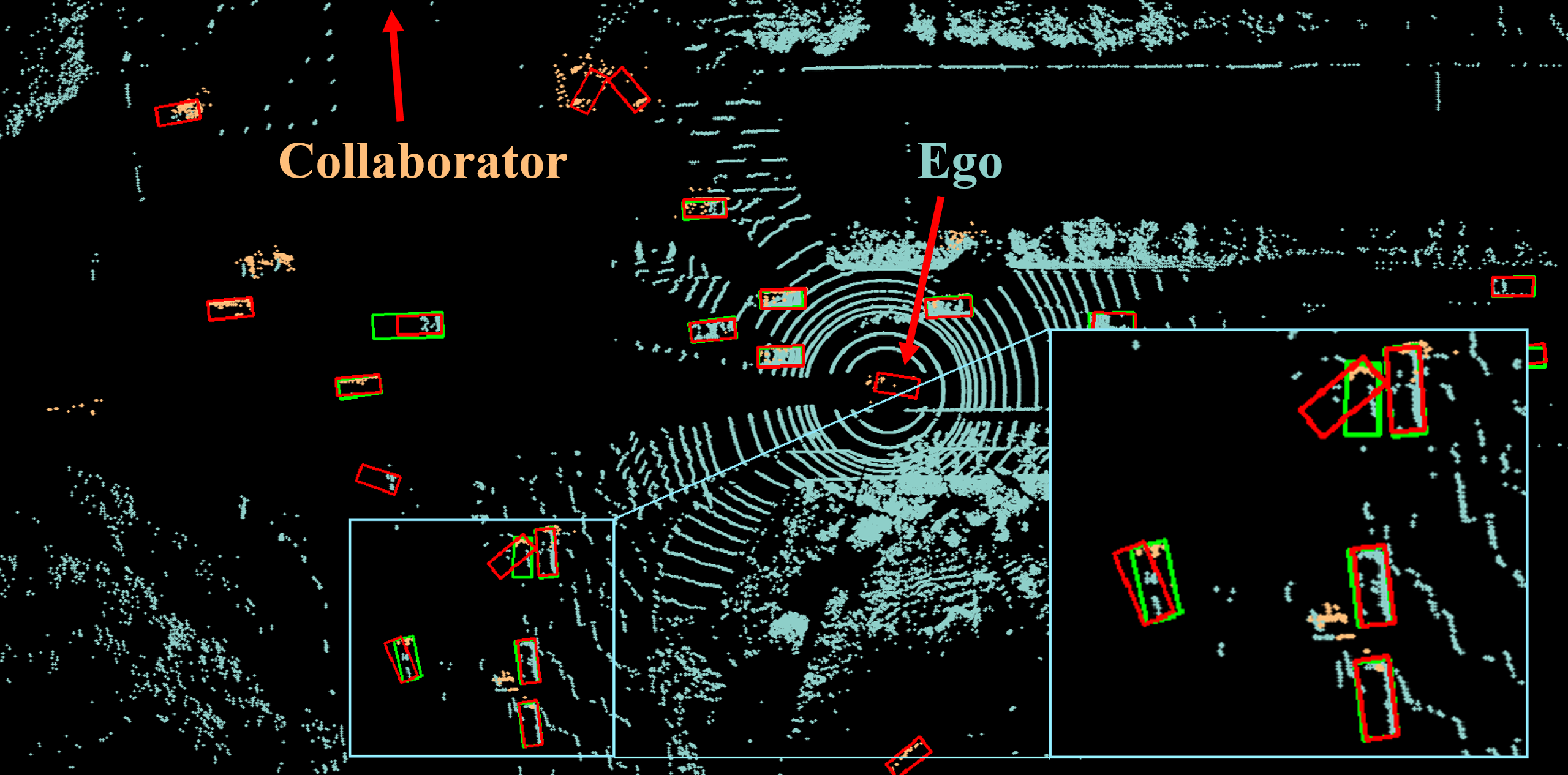}
    \includegraphics[width=0.49\linewidth]{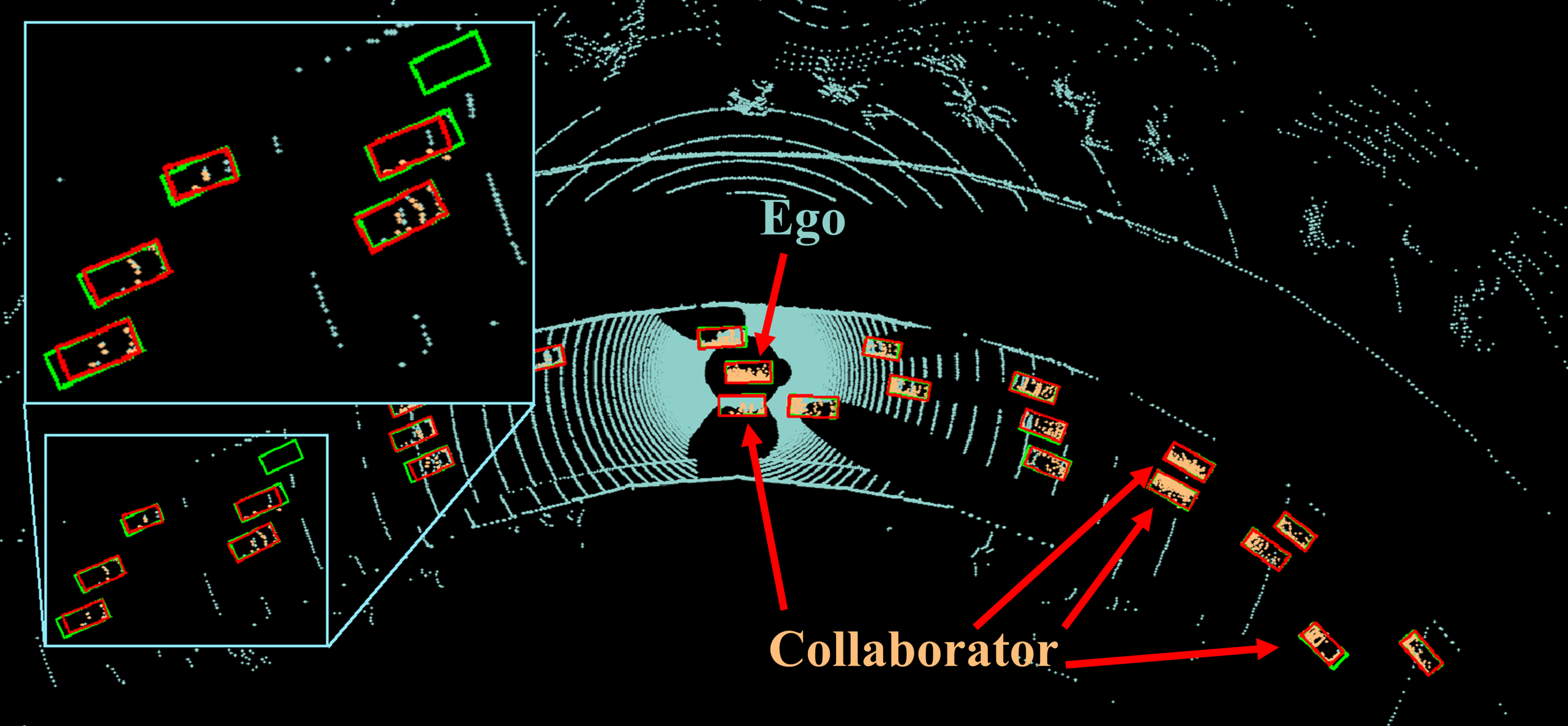}
    \vspace{-1mm}
    \caption{HyComm}
    \vspace{-3mm}
  \end{subfigure}
  \vspace{-3mm}
  \caption{\texttt{HyComm} qualitatively outperforms late, early, intermediate (Where2comm~\cite{hu2022where2comm}) collaboration in DAIR-V2X (left column) and OPV2V (right column). \textcolor{green}{Green} and \textcolor{red}{red} boxes denote ground-truth and detection, respectively. \textcolor[RGB]{173,216,230}{Blue} and \textcolor{orange}{Orange} denote the point clouds from ego and collaborator.}
  \vspace{-6mm}
  \label{Fig:Vis}
\end{figure}

\textbf{Effectiveness of hybrid collaboration.} To validate the effectiveness of hybrid collaboration, Fig.~\ref{Fig:hybrid} compares Late(Box-only), \texttt{HyComm}, and HyComm(Point-only), the latter focusing solely on point messages. We see that~\texttt{HyComm} outperforms HyComm(Point-only) and Late(Box-only), validating box messages and point messages can complement each other and jointly achieve superior performance-bandwidth trade-off. The reason is that: i) point messages enhance performance by rectifying misidentified boxes through additional information, and ii) box messages encode perceptual data more efficiently, thus improving communication efficiency.

\textbf{Effectiveness of confidence-based prioritization.} To validate the effectiveness of confidence-based prioritization, Fig.~\ref{Fig:hybrid} compares HyComm(Point-only) and Point-only. While HyComm(Point-only) emphasizes foreground points, Point-only equally involves background points. We see that HyComm(Point-only) outperforms Point-only in the performance-bandwidth trade-off, underscoring the value of prioritizing foreground perceptual information. The reason is that the foreground perceptual information can help recover the obscured objects in the collaborator's views.

\textbf{Effectiveness of uncertainty-based selection.} To validate the effectiveness of uncertainty-based point selection, Fig.~\ref{Fig:uncertainty} contrasts the performance-bandwidth trade-offs between scenarios with and without uncertainty-based expansion and reweighting. We see that both designs prove effective. The reason is the expansion of imprecise boxes and the assignment of greater weights to points within these expanded boxes allow for more comprehensive coverage of instance-related information, which ultimately improves performance.

\vspace{-4mm}
\subsection{Qualitative results}
\vspace{-1mm}


Fig.~\ref{Fig:Vis} shows the visualization of the detections under DAIR-V2X (left column) and OPV2V (right column) dataset, including single-agent detection (No Collaboration), and late, early, intermediate (Where2comm~\cite{hu2022where2comm}) collaboration methods. The collaboration methods, except for late, are under the same moderate communication budget for a fair comparison. Early collaboration is achieved by randomly sampling points within the budget. Intermediate collaboration is achieved by selecting spatially sparse perceptual features~\cite{hu2022where2comm}. Single-agent detection encounters severe long-range and occlusion issues, resulting in many missed and inaccurate detections, as shown in No Collaboration. All collaboration methods outperform No Collaboration, demonstrating the effectiveness of collaboration. 

Compared to Late, Early, and Where2comm,~\texttt{HyComm} achieves more complete and accurate detection. The reasons are: i) compared to Late which merely merges accurate single-detections and cannot recover the failed detections in single views, by complementing information from various collaborative views,~\texttt{HyComm} can successfully recover these failures; ii) compared to Early,~\texttt{HyComm} prioritizes the more informative foreground information, thereby performing better with the same limited communication budget; and iii) compared to Where2comm which uses uniform message representation,~\texttt{HyComm} offers more flexible representations. Adopting the most compact box messages for precisely encoded targets,~\texttt{HyComm} optimizes bandwidth usage, paving the way for the inclusion of more relevant supplementary information.


\vspace{-2mm}
\section{Conclusion}
\label{sec:conclusion}
\vspace{-2mm}

We propose \texttt{HyComm}, a novel communication-efficient collaborative perception system based on a novel hybrid collaboration strategy, which exchanges both raw observation and perceptual output among agents.
The core idea is to prioritize efficient perceptual output in messages when the communication budget is highly constrained, and then supplement complementary information with raw observation in messages when the communication budget expands.
By doing so, \texttt{HyComm} flexibly adapts to each specific bandwidth requirement across the entire range of communication constraints.
Comprehensive experiments show that \texttt{HyComm} achieves a far superior performance-bandwidth trade-off for LiDAR-based 3D detection.

\vspace{-2mm}



{\small
\bibliographystyle{IEEEtran}
\bibliography{main}
}

\end{document}